%% file: aaai19.tex
\def\year{2019}\relax
\documentclass[letterpaper]{article} 
\usepackage{aaai19}  
\usepackage{times}  
\usepackage{helvet}  
\usepackage{courier}  
\usepackage{url}  
\usepackage{graphicx}  
\usepackage[small]{caption}
\usepackage{xcolor}
\usepackage{soul}
\usepackage{amsmath}
\usepackage{verbatim}
\usepackage{amssymb}
\usepackage{amsthm}
\usepackage{algorithm}
\usepackage{algorithmic}
\usepackage{booktabs}
\usepackage{epstopdf}
\usepackage{lipsum}
\usepackage{color}
\usepackage[normalem]{ulem}
\usepackage{subfigure}
\definecolor{shadecolor}{gray}{0.95}

\newcommand{\csizeten}{
\fontsize{10}{10}\selectfont
}

\newcommand{\tabsize}{
\fontsize{7}{7}\selectfont
}

\definecolor{protein_color}{RGB}{76,153, 0}
\definecolor{catalyst_color}{RGB}{255,128,0}
\definecolor{interaction_color}{RGB}{0,155,155}

\newcommand{\cC}{{\mathcal{C}}}

\newcommand{\cI}{{\mathcal{I}}}

\newcommand{\cH}{{\mathcal{H}}}

\newcommand{\cT}{{\mathcal{T}}}

\newcommand{\bs}[1]{\boldsymbol{#1}}

\newcommand{\thmref}[1]{Theorem~\ref{#1}}
\newcommand{\tabref}[1]{Tab.~\ref{#1}}
\newcommand{\figref}[1]{Fig.~\ref{#1}}
\newcommand{\eqnref}[1]{(Eq. \ref{#1})}
\newcommand{\secref}[1]{Sec.~\ref{#1}}

\newcommand{\algoref}[1]{Alg.~\ref{#1}}

\newtheorem{theorem}{Theorem}

\newtheorem{definition}{Definition}

\begin{document}
%
\title{
Kernelized Hashcode Representations for Relation Extraction
}
%
%
\author{
Sahil Garg$^1$, 
Aram Galstyan$^1$, 
Greg Ver Steeg$^1$,
Irina Rish$^2$, 
Guillermo Cecchi$^2$,
Shuyang Gao$^1$\\
$^1$ USC Information Sciences Institute, Marina del Rey, CA USA\\
$^2$ IBM Thomas J. Watson Research Center, Yorktown Heights, NY USA\\
%
sahil.garg.cs@gmail.com,
\{galstyan, gregv\}@isi.edu, 
\{rish, gcecchi\}@us.ibm.com,
sgao@isi.edu
}
\maketitle
\begin{abstract}
Kernel methods have produced state-of-the-art results for a number of NLP tasks such as relation extraction, but suffer from poor scalability due to the high cost of computing kernel similarities between natural language structures. A recently proposed technique, kernelized locality-sensitive hashing~(KLSH), can significantly reduce the computational cost, but is only applicable to classifiers operating on kNN graphs. Here we propose to use random subspaces of KLSH codes for efficiently constructing an {\em explicit} representation of NLP structures suitable for general classification methods. Further, we propose an approach for optimizing the KLSH model for classification problems by maximizing an approximation of mutual information between the KLSH codes (feature vectors) and the class labels. We evaluate the proposed approach on biomedical relation extraction datasets, and observe significant and robust improvements in accuracy w.r.t. state-of-the-art classifiers, along with drastic (orders-of-magnitude) speedup compared to conventional kernel methods.
\end{abstract}

\section{Introduction}
\begin{figure*}[!tp]
\centering
\includegraphics[
width=1.6\columnwidth]{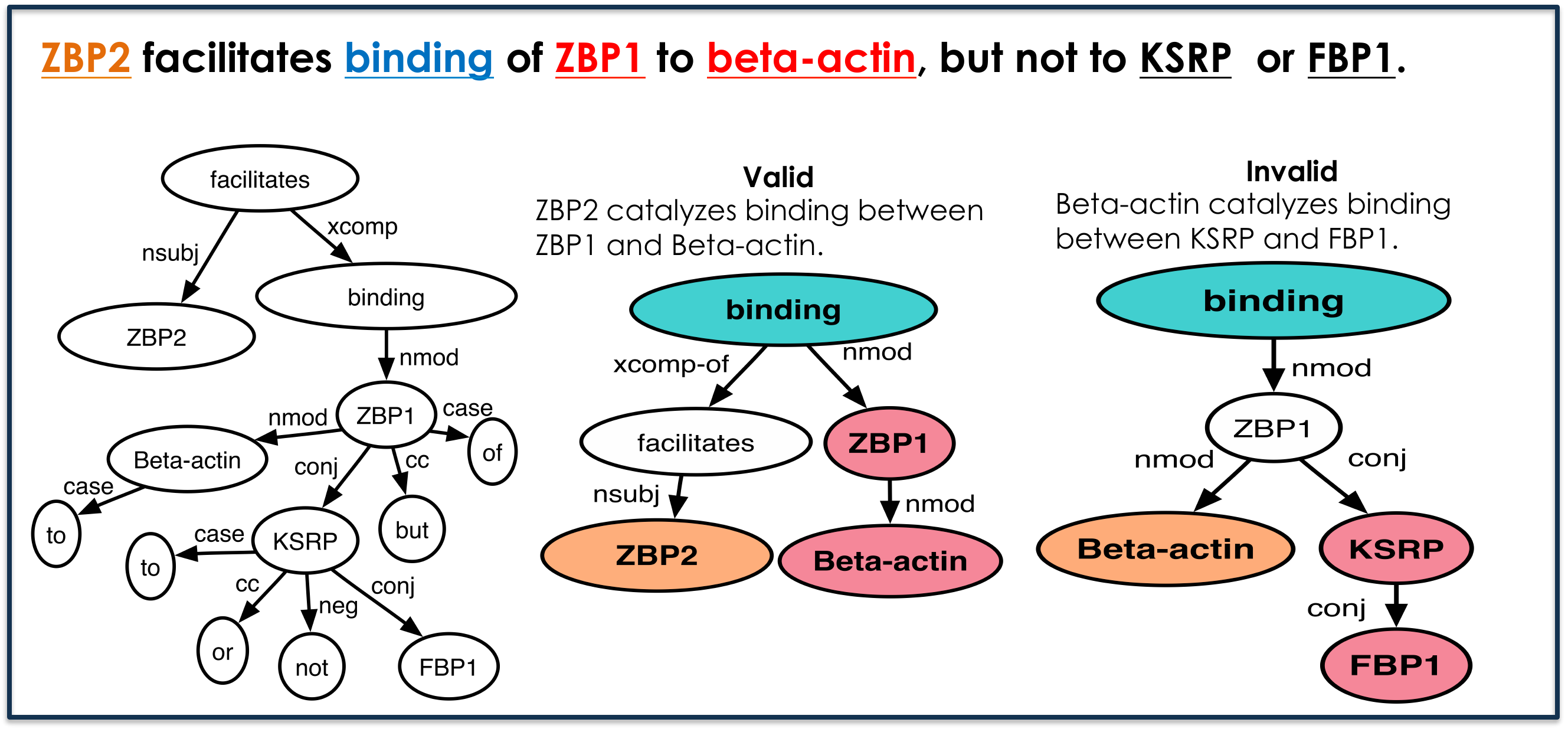}
\caption{On the left, a parse tree of a sentence is shown. In the sentence, the tokens corresponding to  bio-entities~(proteins, chemicals, etc.) or   interaction types are underlined. We highlight the result of  extracting one relation from  the sentence,  using color-coding for its constituents: an interaction type~(green) and bio-entities either participating in the interaction~(red), or catalyzing it~(orange). From two extraction candidates~(valid/invalid), we obtain subgraphs from the parse tree, used as structural features for binary classification of  the candidates.}
\label{fig:sdg_extraction}
\end{figure*}

As the field of biomedical research expands very  rapidly, developing tools for  automated  information extraction from biomedical literature becomes a necessity. In particular, the task of identifying biological entities and their relations from scientific papers has attracted significant   attention  in the past several years~\cite{garg2016extracting,hahn2015domain,krallinger2008overview}, especially because of its potential impact on developing   personalized cancer treatments~\cite{cohen2015darpa,rzhetsky2016big,valenzuelalarge}. See \figref{fig:sdg_extraction} for an example of the relation extraction task.
        
For the relation extraction task, approaches based on convolution kernels~\cite{haussler1999convolution} have demonstrated state-of-the-art performance 
~\cite{chang2016pipe,tikk2010comprehensive}. However, despite their success and intuitive appeal, the traditional kernel-trick based methods can suffer from relatively high computational costs, since computing kernel similarities between two natural language structures~(graphs, paths, sequences, etc.) can be an expensive operation. Furthermore, to build a support vector machine~(SVM) or a k-nearest neighbor~(kNN) classifier from $N$ training examples, one needs to compute kernel similarities between $O(N^2)$ pairs of training points, which can be prohibitively expensive for large $N$. Some approximation methods have been built for scalability of the kernel classifiers. On such approach is kernelized locality-sensitive hashing~(KLSH)~\cite{kulis2009kernelized,joly2011random} that allows  to reduce  the number of kernel computations  to $O(N)$ by providing  efficient approximation for constructing kNN graphs. However, KLSH only works with classifiers that operate on kNN graphs. Thus, {\em the question is whether scalable kernel locality-sensitive hashing approaches can be generalized to a wider range of classifiers}.
    
\begin{figure*}[!tp]
\centering
\includegraphics[
width=1.4\columnwidth]{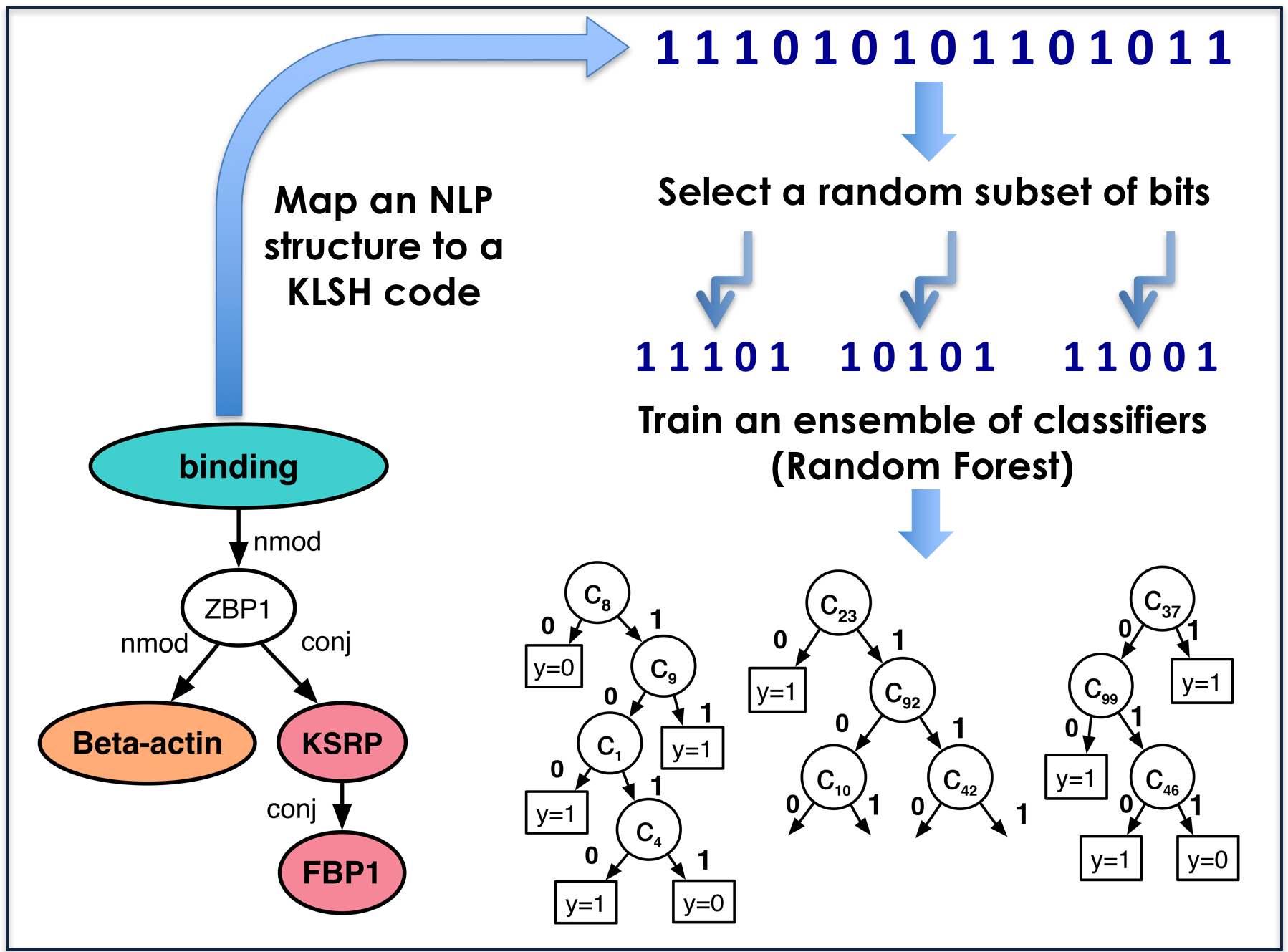}
\caption{On the left, we show  a subgraph from \figref{fig:sdg_extraction} which has to be classified  (we assume binary classification). We map the subgraph to a high-dimensional, kernel similarity-based locality-sensitive hashcode ($\bs{c}$), and use it as a feature vector for  an ensemble classifier. For instance, an efficient and intuitive approach is to  train a Random Forest on binary kernel-hashcodes; in the figure, the nodes in a decision tree makes decisions simply based on hashcode bit values, where  each bit represents  presence or absence of some structural pattern in the subgraph.}
\label{fig:klsh_rf}
\end{figure*}

The main contribution of this paper is  a {\em  principled approach for building   explicit representations for structured data}, as opposed to implicit ones employed in prior  kNN-graph-based approaches, {\em by using random subspaces of KLSH codes.} The intuition behind our approach is as follows. If we keep the total number of bits in the KLSH codes of NLP structures relatively large (e.g., 1000 bits), and take many random subsets of bits~ (e.g., 30 bits each), we can build a large variety of generalized representations corresponding to the subsets, and    preserve detailed  information  present in NLP structures by distributing this information across those   representations.\footnote{Compute cost  of
KLSH codes is linear in the  number of bits~($H$), with the number of kernel computations fixed w.r.t. $H$.}
The main advantage of the {\em proposed representation} is that it  {\em can be used with arbitrary  classification methods},  besides kNN such as, for example,  random forests~(RF)~\cite{ho1995random,breiman2001random}. \figref{fig:klsh_rf} provides   high-level overview of the proposed approach.
	
Our second major contribution is a {\em theoretically justified and computationally efficient method for optimizing  the KLSH representation} with respect to: (1) the kernel function parameters and (2) a reference set of examples w.r.t. which kernel similarities of data samples are computed for obtaining their   KLSH codes. Our approach maximizes (an approximation of) mutual information between KLSH codes of NLP structures and their class labels.
\footnote{See our code here: \url{github.com/sgarg87/HFR}.}

Besides their poor scalability, kernels   usually  involve only a relatively small number of tunable parameters, as opposed to, for instance, neural networks, where the number of parameters can be orders of magnitude larger, thus allowing for more flexible models capable of capturing complex patterns. Our third  important contribution  is a {\em  nonstationary extension of  the conventional convolution kernels}, in order to achieve better expressiveness and flexibility; we achieve this by introducing a  richer parameterization of the kernel similarity function.  Additional parameters, resulting from our non-stationary extension, are also learned by maximizing the mutual information approximation.
    
We validate our model on the relational extraction task using four publicly available datasets. We observe significant improvements in F1 scores w.r.t. the state-of-the-art methods, including recurrent neural nets (RNN), convnets (CNN), and other methods, along with large reductions in the computational complexity as compared to the  traditional kernel-based classifiers. 
    
In summary, our contributions are as follows: (1) we  propose an explicit representation learning for structured data based on kernel locality-sensitive hashing (KLSH), and generalize KLSH-based  approaches in information extraction to work with arbitrary classifiers; (2) we derive an approximation of mutual information and use it for optimizing our models; (3) we increase the expressiveness  of  convolutional kernels by extending their parameterization via a nonstationary extension; (4) we provide an extensive empirical evaluation demonstrating significant advantages of the  approach versus several state-of-art techniques.
 
\section{Background}
\label{sec:oa}

As indicated in \figref{fig:sdg_extraction}, we map the relation extraction task to a classification problem, where each candidate interaction as represented by a corresponding (sub)structure is classified as either valid or invalid. 

Let $\bs{S} = \{ S_i \}_{i=1}^N$ be a set of data points representing NLP structures (such as sequences, paths, graphs) with their corresponding class labels, $\bs{y} = \{ y_i \}_{i=1}^N$. Our goal is to infer the class label of a given test data point $S_*$. Within the kernel-based methods, this is done via a convolution kernel similarity  function $K(S_i,S_j; \bs{\theta})$ defined for any pair of structures $S_i$ and $S_j$ with kernel-parameter $\bs{\theta}$, augmented with an appropriate kernel-based classifier~\cite{garg2016extracting,srivastava2013walk,culotta2004dependency,zelenko2003kernel,haussler1999convolution}.

\subsection{Kernel Locality-Sensitive Hashing~(KLSH)}
\label{sec:klsh_basics}
Previously, Kernel Locality-Sensitive Hashing~(KLSH) was used for constructing approximate kernelized k-Nearest Neighbor (kNN) graphs~\cite{joly2011random,kulis2009kernelized}. The key idea of KLSH as an approximate technique for finding the nearest neighbors of a data point is that rather than computing its similarity w.r.t. all other data points in a given set, the kernel similarity function is computed only w.r.t. the data points in the bucket of its hashcode~(KLSH code). This approximation works well in practice if the hashing approach is locality sensitive, i.e. data points that are very similar to each other are assigned hashcodes with minimal Hamming distance to each other.
	
Herein, we brief on the generic procedure for mapping an arbitrary data point $S_i$  to a binary kernel-hashcode $\bs{c}_i \in \{0, 1\}^{H}$, using a KLSH technique  that relies upon the convolution kernel function $K(.,.; \bs{\theta})$.
	
\begin{figure*}[tp!]
\centering
\includegraphics[width=1.6\columnwidth]{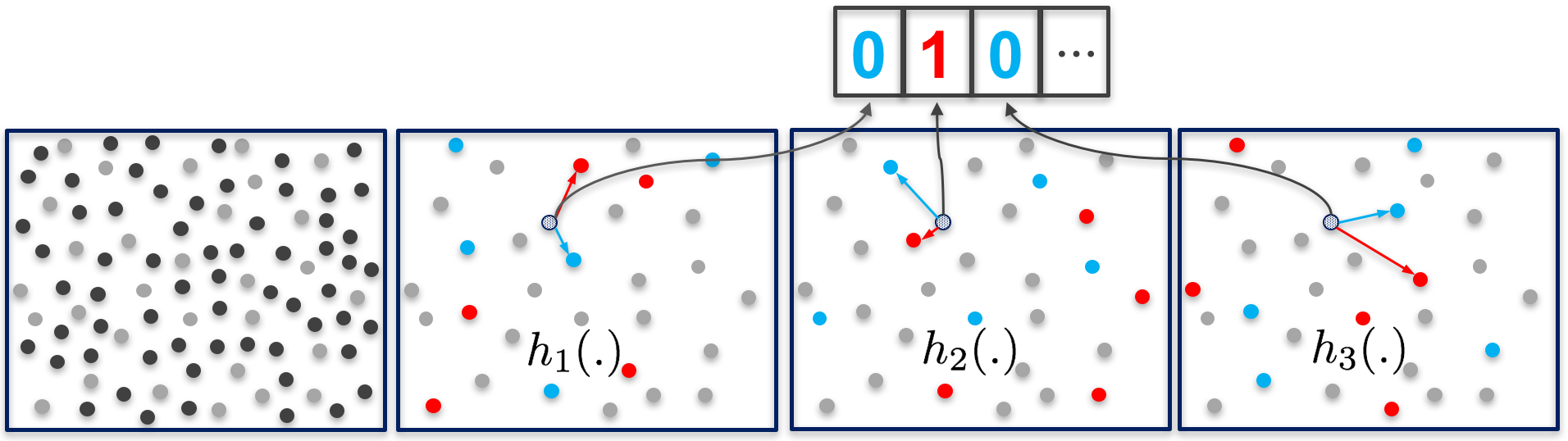}
\caption{
An illustration of the KLSH technique, Random K Nearest Neighbors~(RkNN). First, we obtain a small subset~(gray dots) from a super set of NLP structures as a \emph{reference set} $\bs{S}^R$ that we use for constructing hash functions. For each hash function, two random subsets from the gray dots are obtained, denoted by red and blue. For a given structure, we find its kernel-based 1-nearest neighbor in both of the subsets as indicated by the arrows. Depending on which of the two 1-NNs--either the red 1-NN or the blue 1-NN---is the nearest to the sample, hash function $h_1(.)$ assigns value zero or one to the sample. The same procedure applies to $h_2(.)$ and $h_3(.)$. In this manner, we generate hashcodes with a large number of bits as explicit representations of NLP structures.}
\label{fig:klsh_eg}
\end{figure*}

Let us consider a set of data points $\bs{S}$ that might include both labeled and unlabeled examples. As a first step in constructing the KLSH codes, we select a random subset $\bs{S}^R \subset \bs{S}$ of size $|\bs{S}^R|=M$, which we call a {\em reference set}; this corresponds to the grey dots in the left-most panel of \figref{fig:klsh_eg}. Typically, the size of the reference set is significantly smaller than the size of the whole dataset, $M\ll N$. 

Next, let $\bs{k}_i$ be a real-valued vector of size $M$, whose $j$--th component is the kernel similarity between the data point $S_i$ and the $j$--th element in the reference set, $k_{i,j}=K(S_i,S^R_j; \bs{\theta})$. Further, let $h_l(\bs{k}_i)$, $l=1,\cdots, H$, be a set of $H$ binary valued hash functions that take $\bs{k}_i$ as an input and map it to binary bits and let $\bs{h}(\bs{k}_i)=\{ h_l(\bs{k_i}) \}_{l=1}^H$. The kernel hashcode representation is then given as $\bs{c}_i=\bs{h}(\bs{k}_i)$.
	
We now describe a specific choice of hash functions $h_l(.)$ based on nearest neighbors, called as Random k Nearest Neighbors~(RkNN). For a given $l$, let $\bs{S}_l^{1} \subset \bs{S}^{R} $ and $\bs{S}_l^{2} \subset \bs{S}^{R}$ be two randomly selected, equal-sized and non-overlapping subsets of $\bs{S}^{R}$, $|\bs{S}_l^{1}|=|\bs{S}_l^{2}|=\alpha$, $\bs{S}_l^{1} \cap \bs{S}_l^{2}=\emptyset$. Those sets are indicated by red and blue dots in \figref{fig:klsh_eg}. Furthermore, let $k_{i,l}^1=\max_{S\in\bs{S}_l^{1} }K(S_i, S)$ be the similarity between $S_i$ and its nearest neighbor in $\bs{S}_l^{1}$, with $k_{i,l}^2$ defined similarly (indicated by red and blue arrows in \figref{fig:klsh_eg}). Then the corresponding hash function is:

\begin{equation}
h_l(\bs{k}_i) =
\begin{cases}
1,&\text{if } k_{i,l}^1
< k_{i,l}^2
\\
0,&\text{otherwise}
\end{cases}.
\end{equation}

Pictorial illustration of this hashing scheme is provided in \figref{fig:klsh_eg}, where $S_i$'s nearest neighbors in either subset are indicated by the red and blue arrows.
\footnote{Small value of $\alpha$, i.e. $1 \not \ll \alpha \ll M$, should ensure that hashcode bits have minimal redundancy w.r.t. each other.}
\footnote{In RkNN, since $\alpha \not \gg 1$, $k=1$ should be optimal~\cite{biau2010rate}.}

The same principle of random sub-sampling is applied in KLSH techniques previously proposed in \cite{kulis2009kernelized,joly2011random}. In \cite{joly2011random}, $h_l(.)$ is built by learning a (random) maximum margin boundary~(RMM) that discriminates between the two subsets, $\bs{S}^1_l$ and $\bs{S}^2_l$. In \cite{kulis2009kernelized}, $h_l(.)$ is obtained from $\bs{S}^1_l \cup \bs{S}^2_l$, which is a (approximately) random linear hyperplane in the kernel implied feature space; this is referred to as ``Kulis" here.

In summary, we define klsh$(.; \bs{\theta}, \bs{S}^R)$ as the function, that is parameterized by $\bs{\theta}$ and $\bs{S}^R$, and maps a input data point $S_i$ to its KLSH code $\bs{c}_i$, using the kernel function $K(.,.; \bs{\theta})$ and the set of hash functions $\bs{h}(.)$ as subroutines.

\begin{align}
\bs{c}_i = \text{klsh}(S_i; \bs{\theta}, \bs{S}^R);\ \ 
\bs{c}_i = \bs{h}(\bs{k}_i);\ \ 
k_{i,j} = K(S_i, S^R_j; \bs{\theta})
\label{eqn:klsh_func}
\end{align}

Next, in \secref{sec:klsh_representation_learn}, we propose our approach of learning KLSH codes as generalized representations of NLP structures for classification problems.
    
\section{KLSH for Representation Learning}
\label{sec:klsh_representation_learn}
We propose a novel use of KLSH where the hashcodes~(KLSH codes) can serve as generalized representations (feature vectors) of the data points. Since the KLSH property of being \emph{locality sensitive}~\cite{indyk1998approximate}
\footnote{See a formal definition of locality-sensitive hashing in \cite[Definition 7 in Sec. 4.2]{indyk1998approximate}.}
ensures the data points in the neighborhood of  (or within the same) hashcode bucket are similar, hashcodes should serve as a good representation of the data points.
	
In contrast to the use of KLSH for k-NN, after obtaining the hashcodes for data points, we ignore the step of computing kernel similarities between data points in the neighboring buckets. 
In kNN classifiers using KLSH, a small number of hashcode bits~($H$), corresponding to a small number of hashcode buckets, generate a coarse partition of the feature space---sufficient for approximate computation of a kNN graph. 
In our representation learning framework, however, hashcodes must extract enough information about class labels from the data points, so we propose to generate longer hashcodes, i.e. $H \gg 1$.
It is worthwhile noting that for a \emph{fixed number of kernel computations} $M$ per structure $S_i$~($|\bs{S}^R|=M$), a large number of hashcode bits~($H$) can be generated through the randomization principle with computational cost linear in $H$.
    
Unlike regular kernel methods~(SVM, kNN, etc.), we use kernels to build an explicit feature space, via KLSH. Referring to \figref{fig:klsh_eg}, when using RkNN technique to obtain $\bs{c}_i$ for $S_i$, $l_{th}$ hashcode bit, $\bs{c}_{i,l}$, should correspond to finding a substructure in $S_i$, that should also be present in its 1-NN from either the set $\bs{S}^1_l$ or $\bs{S}^2_l$, depending on the bit value being $0$ or $1$. Thus, $\bs{c}_i$ represents finding important substructures in $S_i$ in relation to $\bs{S}^R$. The same should apply for the other KLSH techniques.
	
\noindent
\subsubsection{Random Subspaces of Kernel Hashcodes:\\}

The next question is how to use the binary-valued representations for building a good classifier.
    
Intuitively, not all the bits may be matching across the hashcodes of NLP structures in training and test datasets; a single classifier learned on all the hashcode bits may overfit to a training dataset. This is especially relevant for bio-information extraction tasks where there is a high possibility of \emph{mismatch between training and test conditions}~\cite{airola2008all,garg2016extracting}; for e.g., in biomedical literature, the mismatch can be due to high diversity of research topics, limited data annotations, variations in writing styles including aspects like hedging, etc. 
So we adopt the approach of building an ensemble of classifiers, with each one built on a random subspace of hashcodes~\cite{zhou2012ensemble,ho1998random}.
    
For building each classifier in an ensemble of $R$ classifiers, $\eta$ bits are selected randomly from $H \gg 1$ hash bits; for inference on a test NLP structure $S_*$, we take mean statistics over the inferred probability vectors from each of the classifiers, as it is a standard practice in ensemble approaches. Another way of building an ensemble from subspaces of hashcodes is bagging~\cite{breiman1996bagging}. If we use a decision tree as a classifier in ensemble, it corresponds to a random forest~\cite{ho1995random,breiman2001random}.
    	
It is highly efficient to train a random forest~(RF) with a large number of decision trees~($R \gg 1$), even on long \emph{binary} hashcodes~($H \gg 1$), leveraging the fact that decision trees can be very efficient to train and test on binary features.

\subsection{Supervised Optimization of KLSH}
\label{sec:opt_ref}

In this section, we propose a framework for optimization of the KLSH codes as generalized representations for a supervised classification task. As described in \secref{sec:klsh_basics}, the mapping of a data points~(an NLP structure $S$) to a KLSH code depends upon the kernel function $K(.,.; \bs{\theta})$ and the reference set $\bs{S}^R$~\eqnref{eqn:klsh_func}. So, within this framework, we optimize the KLSH codes via learning the kernel parameters $\bs{\theta}$, and optionally the reference set $\bs{S}^R$. One important aspect of our optimization setting is that the parameters under optimization are \emph{shared} between all the hash functions jointly, and are not specific to any of the hash functions.
	
\noindent \subsubsection{Mutual Information as an Objective Function:\\}

Intuitively, we want to generate KLSH codes that are maximally informative about the class labels. Thus, for optimizing the KLSH codes, a natural objective function is the mutual information (MI) between KLSH codes of $\bs{S}$ and the class labels, $\cI(\bs{c}:y)$~\cite{cover2012elements}.

\begin{align}
\bs{\theta}^*, {\bs{S}^R}^*
\gets
\mathop{\arg\max}
\limits_{\bs{\theta},\ \bs{S}^R: \bs{S}^R \subset \bs{S}}
\cI(\bs{c}: y);\ 
\bs{c} = \text{klsh}(S; \bs{\theta}, \bs{S}^R)
\label{eqn:corex_opt_thm}
\end{align}

The advantage of MI as the objective, being a fundamental measure of dependence between random variables, is that it is generic enough for optimizing KLSH codes as generalized representations~(feature vectors) of data points to be used with any classifier. Unfortunately, exact estimates of MI function in high-dimensional settings is an extremely difficult problem due to the curse of dimensionality, with the present estimators having very high sample complexity~\cite{kraskov2004estimating,walters2009estimation,singh2014generalized,gao2015efficient,han2015adaptive,wu2016minimax,belghazi18mine}.\footnote{The sample complexity of an entropy estimator for a discrete variable distribution is characterized in terms of its support size $s$, and it is proven to be not less than $s/\log(s)$~\cite{wu2016minimax}. Since the support size for hashcodes is exponential in the number of bits, sample complexity would be prohibitively high unless dependence between the hash code bits is exploited.} Instead, here we \emph{propose to maximize a novel, computationally efficient, good approximation of the MI function}.
    
\noindent \subsubsection{Approximation of Mutual Information:\\}

To derive the approximation, we express the mutual information function as, $\cI(\bs{c}: y) 
= 
\cH(\bs{c}) - \cH(\bs{c}|y)$,
with $\cH(.)$ denoting the Shannon entropy. For binary classification, the expression simplifies to:

\begin{align*}
\cI(\bs{c}: y)
=
\cH(\bs{c})
\!-\! p(y\!=\!0)\cH(\bs{c}|y\!=\!0)
\!-\! p(y\!=\!1) \cH(\bs{c}|y\!=\!1).
\end{align*}

To compute the mutual information, we need to efficiently compute joint entropy of KLSH code bits, $\cH(\bs{c})$. We \emph{propose a good approximation of $\cH(\bs{c})$}, as described below; same applies for $\cH(\bs{c}|y\!=\!0)$ and $\cH(\bs{c}|y\!=\!1)$.

\begin{align}
\cH(\bs{c})
=
\sum_{l=1}^H \cH(c_l) \!-\! \mathcal{TC}(\bs{c})
\approx
\sum_{l=1}^H \cH(c_l) \!-\! \mathcal{TC}(\bs{c};\bs{z^*});
\label{eqn:entropy_approx}
\end{align}

\begin{align}
\cT\cC(\bs{c}; \bs{z}) = \cT\cC(\bs{c}) - \cT\cC(\bs{c}|\bs{z}).
\end{align}

In Eq. \ref{eqn:entropy_approx}, the first term is the sum of marginal entropies for the KLSH code bits. Marginal entropies for binary variables can be computed efficiently. Now, let us understand how to compute the second term in the approximation~\eqnref{eqn:entropy_approx}.
Herein, $\cT\cC(\bs{c}; \bs{z})$ describes the amount of \emph{Total Correlations~(Multi-variate Mutual Information)}~\footnote{``Total correlation" was defined in \cite{watanabe1960information}.} within $\bs{c}$ that can be explained by a latent variables representation $\bs{z}$.

\begin{align}
\bs{z^*}
\gets
\mathop{\arg\max}
\limits_{\bs{z}:|\bs{z}|=|\bs{c}|} \cT\cC(\bs{c}; \bs{z})
\label{eqn:tc_max}
\end{align}

An interesting aspect of the quantity $\mathcal{TC}(\bs{c}; \bs{z})$ is that one can compute it efficiently for optimized $\bs{z^*}$ that explains maximum possible Total Correlations present in $\bs{c}$, s.t. $\mathcal{TC}(\bs{c}| \bs{z}) \approx 0$. In \cite{greg2014discovering}, an unsupervised algorithm called CorEx~\footnote{\url{https://github.com/gregversteeg/CorEx}} is proposed for obtaining such latent variables representation. Their algorithm is efficient for binary input variables, demonstrating a low sample complexity even in very high dimensions of input variables.
Therefore it is particularly relevant for computing the proposed joint entropy approximation on hashcodes. For practical purposes, the dimension of latent representation $\bs{z}$ can be kept much smaller than the dimension of KLSH codes, i.e. $|\bs{z}| \ll H$. This helps to reduce the cost for computing the proposed MI approximation to negligible during the optimization~\eqnref{eqn:corex_opt_thm}.
		
Denoting the joint entropy approximation as $\bar{\cH}(\bs{c})$, we express the approximation of the mutual information as:

\begin{align*}
\bar{\cI}(\bs{c}: y)
=
\bar{\cH}(\bs{c}) 
\!-\! p(y\!=\!0)\bar{\cH}(\bs{c}|y\!=\!0) 
\!-\! p(y\!=\!1) \bar{\cH}(\bs{c}|y\!=\!1).
\label{eqn:mi_approx}
\end{align*}
    
For computation efficiency as well as robustness w.r.t. overfitting, \emph{we use small random subsets~(of size $\gamma$) from a training set for stochastic empirical estimates of $\bar{\cI}(\bs{c}: y)$}, motivated by the idea of stochastic gradients~\cite{bottou2010large}. 
For a slight abuse of notation, when obtaining an empirical estimate of $\bar{\cI}(\bs{c}: y)$ using samples set $\{ \bs{C}, \bs{y} \}$, we simply denote the estimate as: $\bar{\cI}(\bs{C}: \bs{y})$.
Here it is also interesting to note that computation of $\bar{\cI}(\bs{c}: y)$ is very easy to parallelize since the kernel matrices and hash functions can be computed in parallel.

It is worth noting that in our proposed approximation of the MI, both terms need to be computed. In contrast, in the previously proposed variational lower bounds for MI~\cite{barber2003algorithm,chalk2016relevant,alemi2017deep}, MI is expressed as, $\cI(\bs{c}: y) 
= 
\cH(y) - \cH(y|\bs{c})$, so as to obtain a lower bound simply by upper bounding the conditional entropy term with a cross entropy term while ignoring the first term as a constant. Clearly, these approaches are not using MI in its true sense, rather using conditional entropy~(or cross entropy) as the objective. Further, our approximation of MI also allows semi-supervised learning as the first term is computable even for hashcodes of unlabeled examples.

\begin{algorithm*}[tp!]
\caption{Optimizing Reference Set for KLSH}
\begin{algorithmic}[1]
\REQUIRE Train dataset, $\{\bs{S}, \bs{y}\}$;
size of the reference set, $M$; 
$\beta, \gamma$ are number of samples from $\bs{S}$, as candidates for $\bs{S}^R$, and for computing the objective, respectively.
\STATE $\bs{S}^R$ $\gets$ randomSubset($\bs{S}$, $M$)
\label{algline:init_refset}
\COMMENT{$M$ samples from $\bs{S}$
}
\STATE \COMMENT{optimizing the reference set up to size $M$ greedily}
\FOR{$j=1 \to M$}
\label{algline:greedy_opt_refset}
\STATE $\bs{S}^{eo}, \bs{y}^{eo} \gets$ randomSubset($\{\bs{S}, \bs{y}\}, \gamma$)
\label{algline:sample_milb}
\COMMENT{
$\gamma$ samples from $\bs{S}$ for estimating the objective $\bar{\cI}(.)$.
}
\STATE $\bs{S}^{cr}$ $\gets$ randomSubset($\bs{S}$, $\beta$)
\label{algline:sample_candidates}
\COMMENT{$\beta$ samples from $\bs{S}$ as choices for selection to $\bs{S}^R$}
\STATE \COMMENT{iterate over candidates elements for greedy step}
\FOR{$q=1 \to \beta$}
\STATE $S^R_j \gets S^{cr}_q$ \COMMENT{$S^{cr}_q$ is a choice for selection to $\bs{S}^R$}
\STATE $\bs{c}^{eo}_i \gets$ klsh($S^{eo}_i$; $\bs{\theta}$, $\bs{S}^R$) $\forall{S^{eo}_i \in \bs{S}^{eo}}$
\COMMENT{Eq. \ref{eqn:klsh_func}}
\STATE $\bs{mi}_q \gets$ $\bar{\cI}$($\bs{C}^{eo}, \bs{y}^{eo}$)
\label{algline:milb}
\COMMENT{estimating objective}
\ENDFOR
\STATE $S^R_j \gets$ chooseElementWithMaxMI($\bs{mi}$, $\bs{S}^{cr}$)
\label{algline:max_mi_lb}
\ENDFOR
\RETURN $\bs{S}^R$
\end{algorithmic}
\label{alg:opt_ref}
\end{algorithm*}
    
\subsubsection{Algorithms for Optimization:\\}
Using the proposed approximate mutual information function as an objective, one can optimize the kernel parameters either using grid search or an MCMC procedure.

For optimizing the reference set $\bs{S}^R$~(of size $M$) as a subset of $\bs{S}$, via maximization of the same objective, we propose a greedy algorithm with pseudo code in \algoref{alg:opt_ref}. Initially, $\bs{S}^R$ is initialized with a random subset of $\bs{S}$~(line \ref{algline:init_refset}). Thereafter, $\bar{\cI}(.)$ is maximized greedily, updating one element in $\bs{S}^R$ in each greedy step~(line \ref{algline:greedy_opt_refset}); greedy maximization of MI-like objectives has been successful~\cite{gao2016variational,krause2008near}.
Employing the paradigm of stochastic sampling, 
for estimating $\bar{\cI}(.)$, we randomly sample a small subset of $\bs{S}$~(of size $\gamma$) along with their class labels~(line \ref{algline:sample_milb}). Also, in a single greedy step, we consider only a small random subset of $\bs{S}$~(of size $\beta$) as candidates for selection into $\bs{S}^R$~(line \ref{algline:sample_candidates}); for $\beta \gg 1$, with high probability, each element in $\bs{S}$ should be seen as a candidate at least once by the algorithm. \algoref{alg:opt_ref} requires kernel computations of order, $O(\gamma M^2+\gamma\beta M)$, with $\beta, \gamma$ being the sampling size constants; in practice, $M \ll N$. Note that $\bs{\theta}$ and $\bs{S}^R$ can be optimized in an iterative manner.
 
\subsection{Nonstationary Extension for Kernels}
\label{sec:nst}

One common principle applicable to all the convolution kernel functions, $K(.,.)$, defining similarity between any two NLP structures is: \emph{$K(., .)$ is expressed in terms of a kernel function, $k(.,.)$}, that defines similarity between any two tokens~(node/edge labels in \figref{fig:sdg_extraction}). Some common examples of $k(.,.)$, from previous works~\cite{culotta2004dependency,srivastava2013walk}, are:

\begin{align*}
\text{Gaussian: } 
k(a,b) = \exp(-||\bs{w}_a - \bs{w}_j||_b^2),
\\
\text{Sigmoid: }
k(a,b) = (1 + \tanh(\bs{w}_a^T\bs{w}_b))/2.
\end{align*}
	
\noindent Herein, $a, b$ are tokens, and $\bs{w}_a$, $\bs{w}_b$ are the corresponding word vectors. The first kernels is stationary, i.e. translation invariant~\cite{genton2001classes}, and the second one is nonstationary, although lacking nonstationarity-specific parameters for learning nonstationarity in a data-driven manner.
		
There are generic nonstationarity-based parameterizations, unexplored in NLP, applicable for extending any kernel, $k(.,.)$, to a nonstationary one, $k^{NS}(.,.)$, so as to achieve higher \emph{expressiveness and generalization} in model learning~\cite{paciorek2003nonstationary,rasmussen2006gaussian}. For NLP, nonstationarity of $K(.,.)$ can be formalized as in \thmref{thm:nst_nlp}; see the longer version of this paper for a proof.

\begin{theorem}
A convolution kernel $K(.,.)$, that is a function of the kernel $k(.,.)$, is stationary if $k(.,.)$ is stationary, and vice versa. From a non-stationary $k^{NS}(.,.)$, the corresponding extension of $K(.,.)$, $K^{NS}(.,.)$, is also guaranteed to be a valid non-stationary convolution kernel.
\label{thm:nst_nlp}
\end{theorem}

One simple and intuitive nonstationary extension of $k(.,.)$ is: $k^{NS}(a,b) = 
\sigma_a k(a,b) \sigma_b$.
\noindent Here, $\sigma \geq 0$, are nonstationarity-based parameters; for more details, see \cite{rasmussen2006gaussian}; another choice for the nonstationary extension is based on the concept of process convolution, as proposed in \cite{paciorek2003nonstationary}.
If $\sigma_a=0$, it means that the token $a$ should be completely ignored when computing a convolution kernel similarity of an NLP structure~(tree, path, etc.) that contains the token $a$ (node or edge label $a$) w.r.t. another NLP structure. Thus, the additional nonstationary parameters allow convolution kernels to be expressive enough for deciding if some substructures in an NLP structure should be ignored explicitly.\footnote{This approach is explicit in ignoring sub-structures irrelevant for a given task unlike the~(complementary) standard skipping over non-matching substructures in a convolution kernel.}

\begin{table*}[tp!]
\centering
%
\renewcommand{\tabcolsep}{17pt}
\begin{tabular}{lll}
		\toprule
		\textbf{Models}
&\textbf{(AIMed, BioInfer)}&\textbf{(BioInfer, AIMed)}\\
        \toprule
SVM$_1$~\cite{airola2008all}
&0.25&0.44\\
%
SVM$_2$~\cite{airola2008all}
&0.47&0.47\\
\midrule
SVM~\cite{miwa2009protein}
&0.53&0.50\\
\midrule
SVM~\cite{tikk2010comprehensive}
&0.41&0.42\\
&(0.67, 0.29)&(0.27, 0.87)\\
\toprule
CNN~\cite{nguyen2015relation}
&0.37&0.45\\
\midrule
Bi-LSTM~\cite{kavuluru2017extracting}
&0.30&0.47\\
\midrule
CNN~\cite{peng2017deep}
&0.48&0.50\\
&(0.40, 0.61)&(0.40, 0.66)\\
\midrule
RNN~\cite{hsieh2017identifying}
&0.49&0.51\\
%
%
%
%
\midrule
CNN-RevGrad~\cite{ganin2016domain}
&0.43&0.47\\
\midrule
Bi-LSTM-RevGrad~\cite{ganin2016domain}
&0.40&0.46\\
\midrule
Adv-CNN~\cite{rios2018generalizing}
&0.54&0.49\\
\midrule
Adv-Bi-LSTM~\cite{rios2018generalizing}
&$\bs{0.57}$&0.49\\
\midrule
KLSH-kNN&$0.51$&$0.51$\\
&(0.41, 0.68)&(0.38, 0.80)\\
%
%
\toprule
%
%
%
\textbf{KLSH-RF}
&$\bs{0.57}$&$\bs{0.54}$\\
&\textbf{(0.46, 0.75)}&\textbf{(0.37, 0.95)}\\
\toprule
\end{tabular}
\caption{Cross-corpus evaluation results for (training, test) pairs of PPI datasets, AIMed and BioInfer datasets. For each model, we report F1 score in the first row corresponding to it. In some of the previous works, precision, recall numbers are not reported; wherever available, we show precision, recall numbers as well, in brackets. 
}
\vspace{-3mm}
\label{tab:results_ppi}
\end{table*}

While the above proposed idea of nonstationary kernel extensions for NLP structures remains general, for the experiments, the nonstationary kernel for similarity between tuples with format (edge-label, node-label) is defined as the product of kernels on edge labels, $e_a, e_b$, and node labels, $n_a, n_b$,

\begin{equation*}
k^{NS}
\!
((e_i, n_i),\! (e_j, n_j))
= 
\sigma_{e_i}
k_e(e_i, e_j)
\sigma_{e_j}
k_n(n_i, n_j),
\end{equation*}

with $\sigma$ operating only on edge labels. Edge labels come from syntactic or semantic parses of text with small size vocabulary~(see syntactic parse-based edge labels in \figref{fig:sdg_extraction}); we keep $\sigma \in \{ 0, 1 \}$ as a measure for robustness to over-fitting. These parameters are learned by maximizing the same objective, $\bar{\cI}(.)$, using the well known Metropolis-Hastings MCMC procedure~\cite{hastings1970monte}.

\section{Experiments}
\label{sec:exp}

We evaluate our model ``KLSH-RF"~(kernelized locality-sensitive hashing with random forest) for the biomedical relation extraction task using four public datasets, AIMed, BioInfer, PubMed45, BioNLP, as briefed below.\footnote{PubMed45 dataset is available here: \url{github.com/sgarg87/big_mech_isi_gg/tree/master/pubmed45_dataset}; the other three datasets are here: \url{corpora.informatik.hu-berlin.de}} \figref{fig:sdg_extraction} illustrates that the task is formulated as a binary classification of extraction candidates. For evaluation, it is standard practice to compute precision, recall, and F1 score on the positive class~(i.e., identifying valid extractions).

\noindent\subsubsection{Details on Datasets and Structural Features:\\}

\noindent\emph{AIMed and BioInfer:} For AIMed and BioInfer datasets, cross-corpus evaluation has been performed in many previous works~\cite{airola2008all,tikk2010comprehensive,peng2017deep,hsieh2017identifying}. Herein, the task is of identifying pairs of interacting proteins~(PPI) in a sentence while ignoring the interaction type. 
We follow the same evaluation setup, using Stanford Dependency Graph parses of text sentences to obtain undirected shortest paths as structural features for use with a path kernel~(PK) 
to classify protein-protein pairs.

\noindent\emph{PubMed45 \& BioNLP:}
We use PubMed45 and BioNLP datasets for an extensive evaluation of our KLSH-RF model; for more details on the two datasets, see \cite{garg2016extracting} and \cite{kim2009overview,kim2011overview,nedellec2013overview}. Annotations in these datasets are richer in the sense that a bio-molecular interaction can involve up to two participants, along with an optional catalyst, and an interaction type from an unrestricted list. In PubMed45~(BioNLP) dataset, 36\%~(17\%) of the ``valid" interactions are such that an interaction must involve two participants and a catalyst. For both datasets, we use abstract meaning representation~(AMR) to build subgraph or shortest path-based structural features~\cite{Banarescu13abstractmeaning}, for use with graph kernels~(GK) or path kernels~(PK) respectively, as done in the recent works evaluating these datasets~\cite{garg2016extracting,raobiomedical}.
For a fair comparison of the classification models, we use the same bio-AMR parser~\cite{pust2015parsing} as in the previous works. In \cite{garg2016extracting}, the PubMed45 dataset is split into 11 subsets for evaluation, at paper level. Keeping one of the subsets for testing, we use the others for training a binary classifier. This procedure is repeated for all 11 subsets in order to obtain the final F1 scores~(mean and standard deviation values are reported from the numbers for 11 subsets). For BioNLP dataset~\cite{kim2009overview,kim2011overview,nedellec2013overview}, we use training datasets from years 2009, 2011, 2013 for learning a model, and the development dataset from year 2013 as the test set; the same evaluation setup is followed in \cite{raobiomedical}.

\begin{table*}[tp!]
\centering
%
\renewcommand{\tabcolsep}{17pt}
\begin{tabular}{llll}
\toprule
\textbf{Models}&\textbf{PubMed45}&\textbf{PubMed45-ERN}&\textbf{BioNLP}\\
\toprule
SVM~\cite{garg2016extracting}&$0.45$$\pm$$0.25$&$0.33$$\pm$$0.16$&$0.46$\\
&(0.58, 0.43)&(0.33, 0.45)&(0.35, 0.67)\\
\toprule
LSTM~\cite{raobiomedical}&N.A.&N.A.&0.46\\
&&&(0.51, 0.44)\\
\midrule
LSTM
&$0.30$$\pm$$0.21$
&$0.29$$\pm$$0.14$
&0.59\\
&(0.38, 0.28)
&(0.42, 0.33)
&(0.89, 0.44)\\
\midrule
Bi-LSTM
&$0.46$$\pm$$0.26$
&$0.37$$\pm$$0.15$
&0.55\\
&(0.59, 0.43)
&(0.45, 0.40)
&(0.92, 0.39)\\
\midrule
LSTM-CNN
&$0.50$$\pm$$0.27$
&$0.31$$\pm$$0.17$
&0.60\\
&(0.55, 0.50)
&(0.35, 0.40)
&(0.77, 0.49)\\
\midrule
CNN
&$0.51$$\pm$$0.28$
&$0.33$$\pm$$0.18$
&0.60\\
&(0.46, 0.46)
&(0.36, 0.32)
&(0.80, 0.48)\\
\midrule
KLSH-kNN&$0.46$$\pm$$0.21$&$0.23$$\pm$$0.13$&$0.60$\\
&(0.44, 0.53)&(0.23, 0.29)&(0.63, 0.57)\\
\toprule
\textbf{KLSH-RF}&$\bs{0.57}$$\pm$$\bs{0.25}$&$\bs{0.45}$$\pm$$\bs{0.22}$
&$\bs{0.63}$\\
&(\textbf{0.63, 0.55})&(\textbf{0.51, 0.52})
&(\textbf{0.78, 0.53})\\
\toprule
\end{tabular}
\caption{Evaluation results for PubMed45  and BioNLP datasets. For each model, we report F1 score~(mean $\pm$ standard deviation) in the first row corresponding to it, and show mean-precision, mean-recall numbers as well, in brackets. For BioNLP, we don't show standard deviation since there is only one fixed test subset.}
\label{tab:results_pubmed45_bionlp_all}
\end{table*}
    
In addition to the models previously evaluated on these datasets, we also compare our KLSH-RF model to KLSH-kNN~(kNN classifier with KSLH approximation).

For PubMed45 and BioNLP datasets, for the lack of evaluations of previous works on these datasets, we perform extensive empirical evaluation ourselves of competitive neural network models, LSTM, Bi-LSTM, LSTM-CNN, CNN; from fine-grained tuning, for PubMed45 \& PubMed45-ERN datasets, the tuned neural architecture was a five-layer network, [8, 16, 32, 16, 8], having 8, 16,
32, 16, and 8 nodes, respectively, in the 1st, 2nd, 3rd, 4th, 5th hidden
layers; for BioNLP dataset, the tuned neural architecture was a two layer network, [32, 32].

\noindent\subsubsection{Parameter Settings:\\}

We use GK and PK, both using the same word vectors, with kernel parameter settings same as in \cite{garg2016extracting,mooney2005subsequence}.

Reference set size, $M$, doesn't need tuning in our proposed model; there is a trade-off between compute cost and accuracy; by default, we keep $M=100$. For tuning any other parameters in our model or competitive models, including the choice of a kernel similarity function~(PK or GK), we use 10\% of training data, sampled randomly, for validation purposes. From a preliminary tuning, we set parameters, $H=1000$, $R=250$, $\eta=30$, $\alpha=2$, and choose RMM as the KLSH technique from the three choices discussed in \secref{sec:klsh_basics}; same parameter values are used across all the experiments unless mentioned otherwise.
    
When selecting reference set $\bs{S}^R$ randomly, we perform 10 trials, and report mean statistics. (Variance across these trials is small, empirically.) The same applies for KLSH-kNN. When optimizing $\bs{S}^R$ with \algoref{alg:opt_ref}, we use $\beta$$=$$1000$, $\gamma$$=$$300$~(sampling parameters are easy to tune). We employ 4 cores on an i7 processor, with 16GB memory.
	
\subsection{Main Results for KLSH-RF}

In the following we compare the simplest version of our KLSH-RF model that is optimized by learning the kernel parameters via maximization of the MI approximation, as described in \secref{sec:opt_ref}~($\gamma=1000$). In summary, our KLSH-RF model outperforms state-of-the-art models consistently across the four datasets, along with very significant speedups in training time w.r.t. traditional kernel classifiers.
	
\noindent\subsubsection{Results for AIMed and BioInfer Datasets:\\}

In reference to \tabref{tab:results_ppi}, KLSH-RF gives an F1 score significantly higher than state-of-the-art kernel-based models~(6 pts gain in F1 score w.r.t. KLSH-kNN), and consistently outperforms the neural models. When using AIMed for training and BioInfer for testing, there is a tie between Adv-Bi-LSTM~\cite{rios2018generalizing} and KLSH-RF. However, KLSH-RF still outperforms their Adv-CNN model by 3 pts; further, the performance of Adv-CNN and Adv-Bi-LSTM is not consistent, giving a low F1 score when training on the BioInfer dataset for testing on AIMed. For the latter setting of AIMed as a test set, we obtain an F1 score improvement by 3 pts w.r.t. the best competitive models, RNN \& KLSH-kNN. Overall, the performance of KLSH-RF is more consistent across the two evaluation settings, in comparison to any other competitive model.
    
The models based on adversarial neural networks~\cite{ganin2016domain,rios2018generalizing}, Adv-CNN, Adv-Bi-LSTM, CNN-RevGrad, Bi-LSTM-RevGrad, are learned jointly on labeled training datasets and unlabeled test sets, whereas our model is purely supervised. In contrast to our principled approach, there are also system-level solutions using multiple parses jointly, along with multiple kernels, and knowledge bases~\cite{miwa2009protein,chang2016pipe}. We refrain from comparing KLSH-RF w.r.t. such system level solutions, as it would be an unfair comparison from a modeling perspective.
    
\noindent\subsubsection{Results for PubMed45 and BioNLP Datasets:\\}

A summary of main results is presented in \tabref{tab:results_pubmed45_bionlp_all}. ``PubMed45-ERN" is another version of the PubMed45 dataset from \cite{garg2016extracting}, with ERN referring to entity recognition noise. Clearly, our model gives F1 scores significantly higher than SVM, LSTM, Bi-LSTM, LSTM-CNN, CNN, and KLSH-kNN model. For PubMed45, PubMed45-ERN, and BioNLP, the F1 score for KLSH-RF is higher by 6 pts, 8 pts, and 3 pts respectively w.r.t. state of the art; KLSH-RF is the most consistent in its performance across the datasets and significantly more scalable than SVM. Note that standard deviations of F1 scores are high for the PubMed45 dataset~(and PubMed45-ERN) because of the high variation in distribution of text across the 11 test subsets~(the F1 score improvements with our model are statistically significant, p-value=4.4e-8).
	
For the PubMed45 dataset, there are no previously published results with a neural model~(LSTM). The LSTM model of \cite{raobiomedical}, proposed specifically for the BioNLP dataset, is not directly applicable for the PubMed45 dataset because the list of interaction types in the latter is unrestricted. F1 score numbers for SVM classifier were  also improved in \cite{garg2016extracting} by additional contributions such as  document-level inference, and the joint use of semantic and syntactic representations; those system-level contributions are complementary to ours, so excluded from the comparison.
	
\subsection{Detailed Analysis of KLSH-RF}

\begin{figure*}[tp!]
\centering
\subfigure[
Reference Set Opt.~(PubMed45)
]{
\includegraphics[
width=0.7\columnwidth]{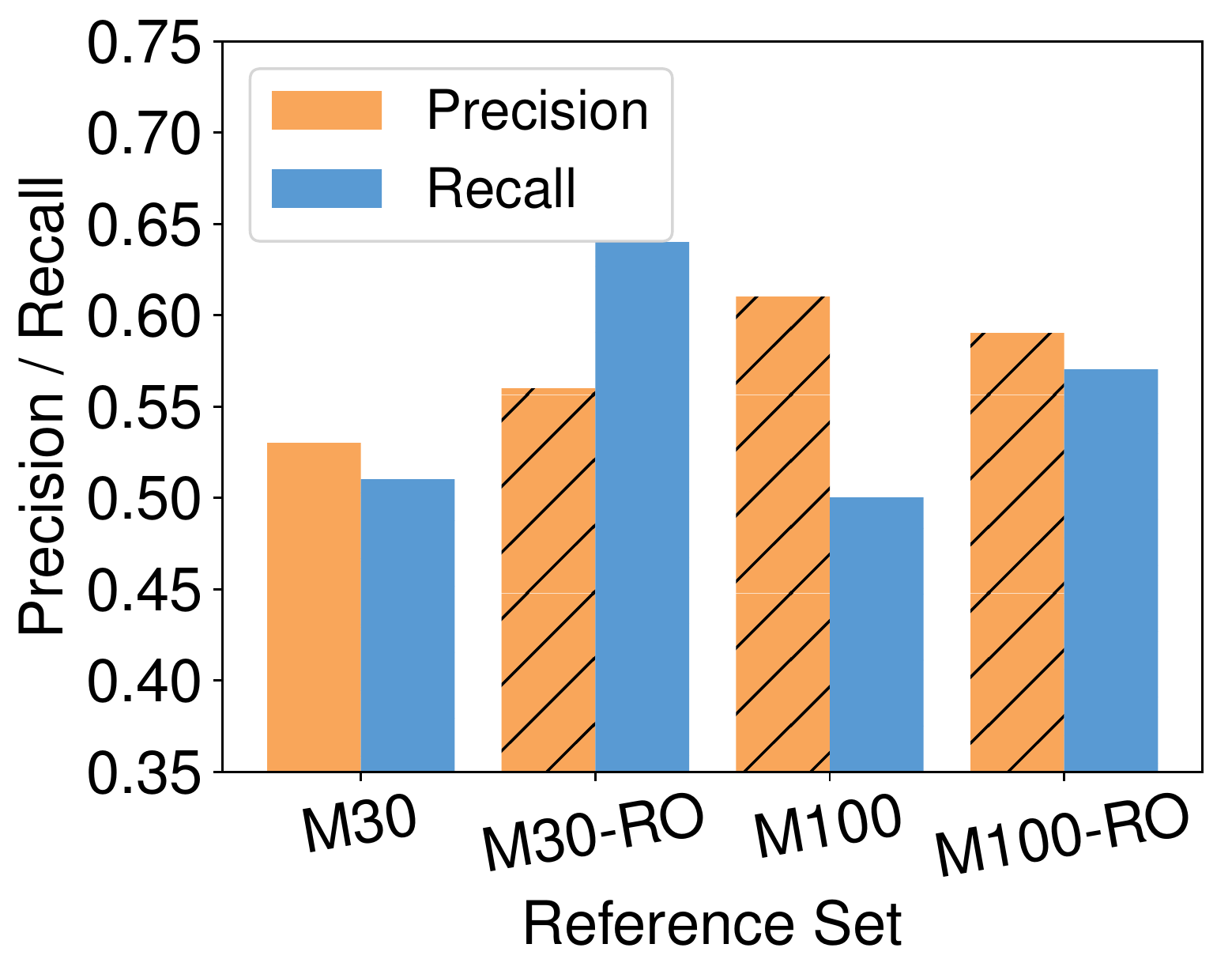}
\label{fig:pubmed_ro}
}
%
%
\subfigure[
Reference Set Opt.~(BioNLP)
]{
\includegraphics[
width=0.7\columnwidth
]{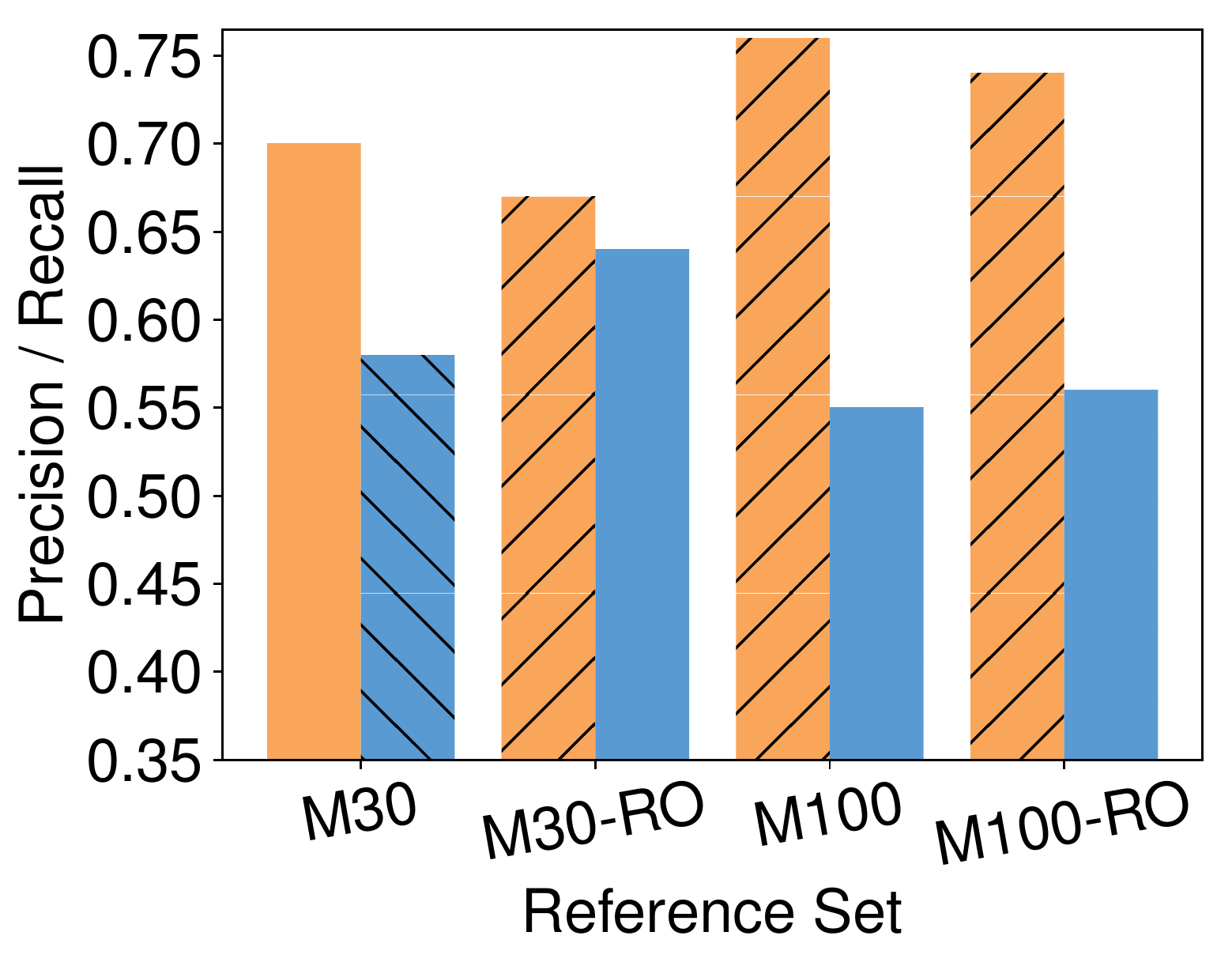}
\label{fig:bionlp_ro}
}
%
%
\subfigure[
NSK Learning~(PubMed45)
]{
\includegraphics[
width=0.7\columnwidth]{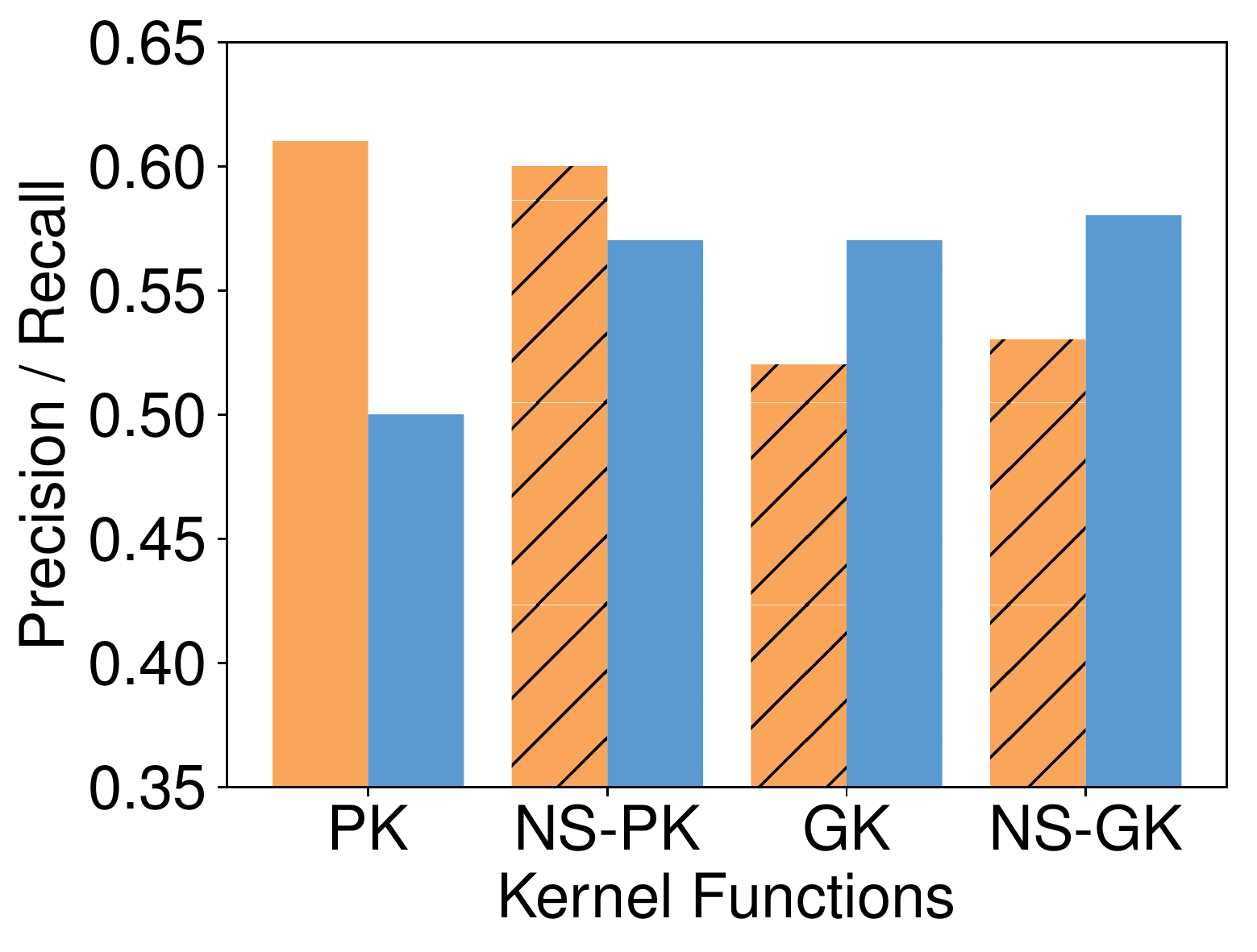}
\label{fig:pubmed_ns}
}
%
%
\subfigure[
NSK Learning~(BioNLP)
]{
\includegraphics[
width=0.7\columnwidth
]{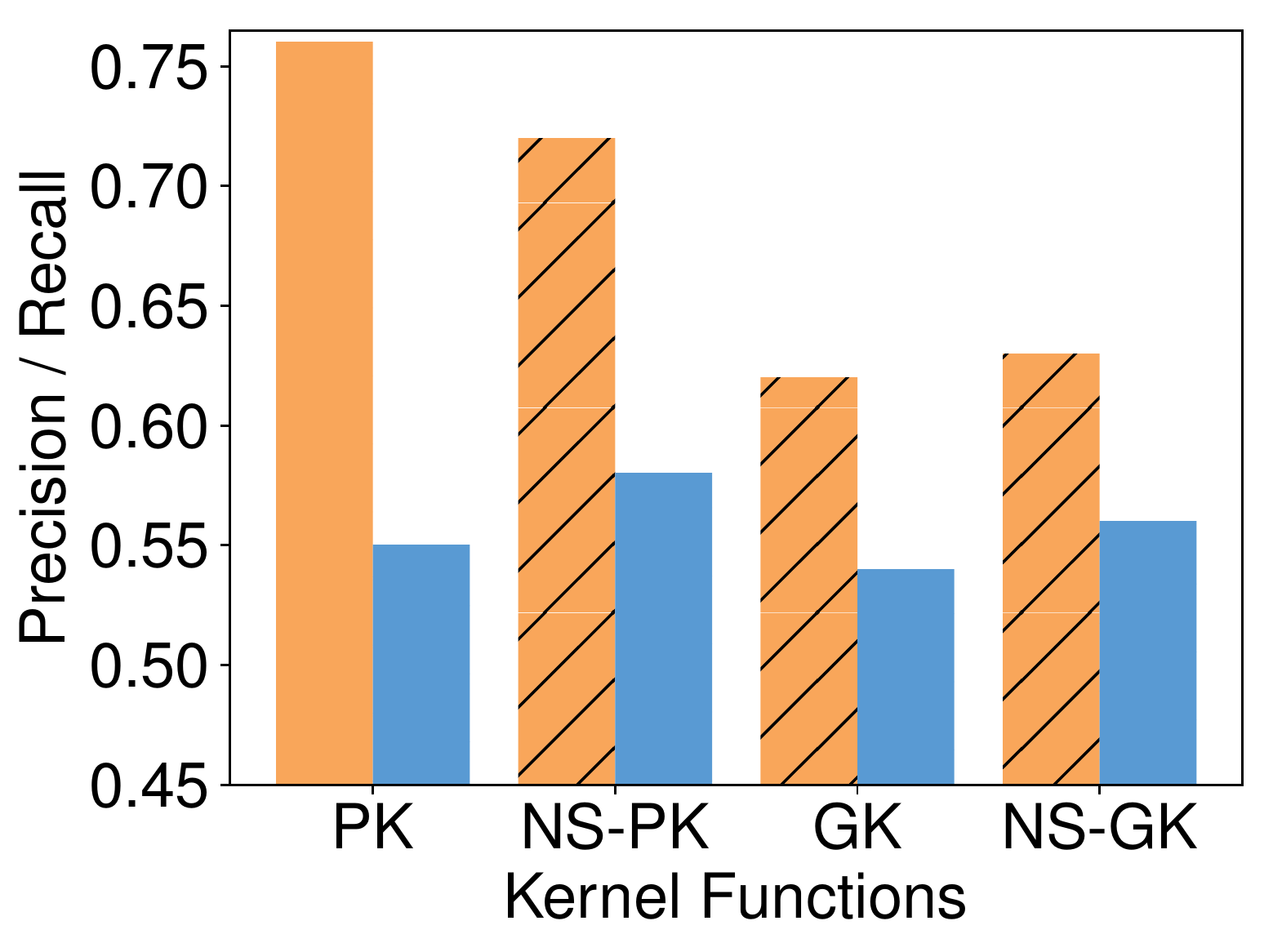}
\label{fig:bionlp_ns}
}
%
%
\subfigure[
\tabsize
Training Time~(BioNLP)
]{
\includegraphics[
width=0.75\columnwidth
]{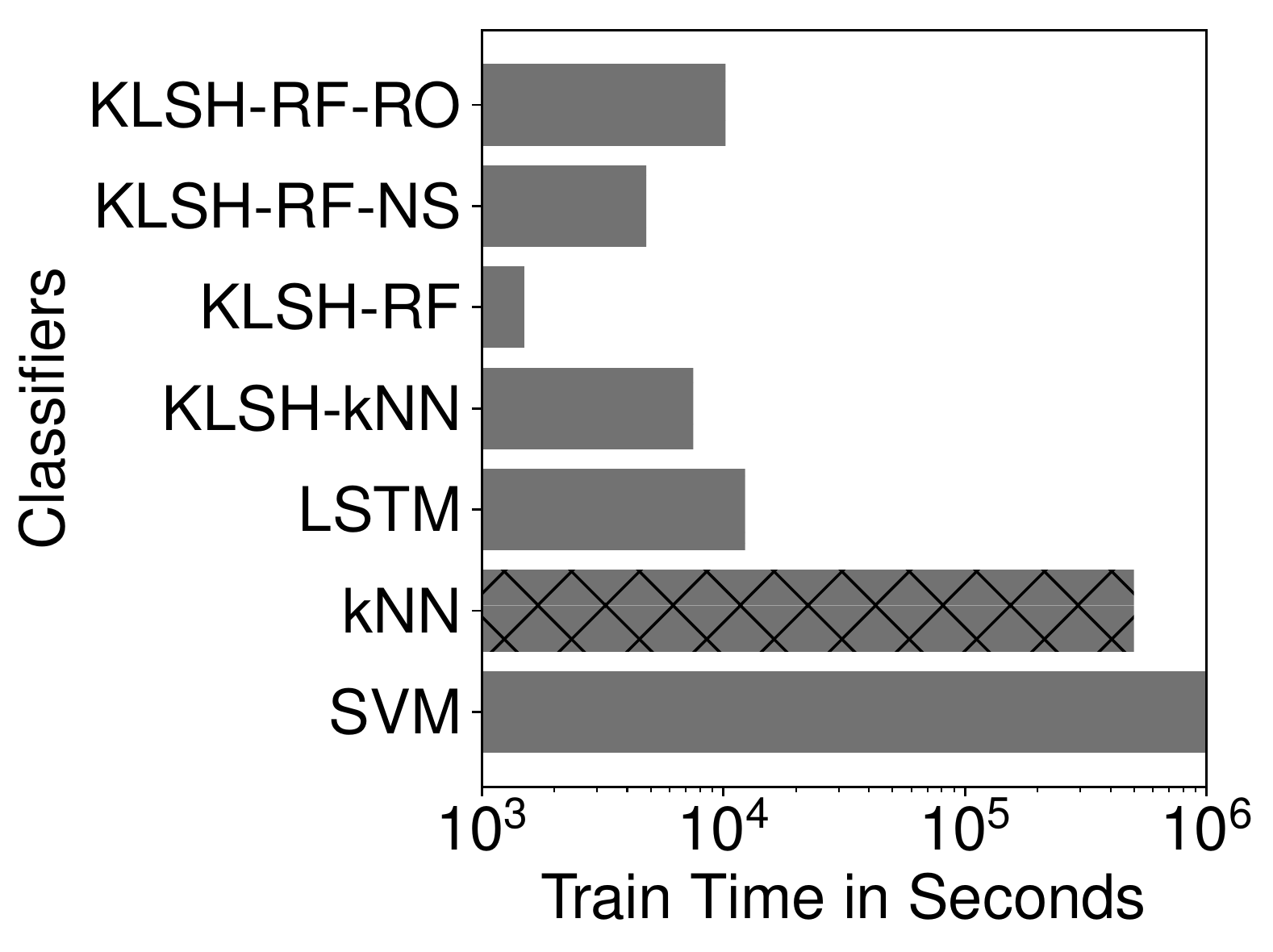}
\label{fig:compute_time}
}
\caption{Detailed Evaluation of KLSH-RF model, using PubMed45 and BioNLP datasets. Here, orange and blue bars are for precision and recall numbers respectively. ``NSK" refers to nonstationary kernel learning; PK
\& GK denote Path Kernels and Graph Kernels respectively; NS-PK and NS-GK are extensions of PK and GK respectively, with addition of nonstationarity based binary parameters; ``M30" represents $\bs{S}^R$ of size $30$ selected randomly, and the suffix ``RO" in ``M30-RO" refers to optimization of $\bs{S}^R$~(Reference optimization) in contrast to random selection of $\bs{S}^R$.}
\label{fig:klsh_rf_opt_details}
\end{figure*}

While we are able to obtain superior results with our basic KLSH-RF model w.r.t. state-of-the-art methods using just core optimization of the kernel parameters $\bs{\theta}$, in this subsection we analyze how we can further improve the model.
In \figref{fig:klsh_rf_opt_details} we present our results from optimization of other aspects of the KLSH-RF model: reference set optimization~(RO) and non-stationary kernel parameters learning~(NS).
(In the longer version of this paper, we also analyze the effect of parameters, $H, R$, and the choice of KLSH technique, under controller experiment settings.) We report mean values for precision, recall, F1 scores. For these experiments, we focus on PubMed45 and BioNLP datasets.
    
\noindent\subsubsection{Reference Set Optimization:}

In \figref{fig:pubmed_ro} and \ref{fig:bionlp_ro}, we analyze the effect of the reference set optimization~(RO), in comparison to random selection, and find that the optimization leads to significant increase in recall~(7-13 pts) for PubMed dataset along with a marginal increase/decrease in precision~(2-3 pts); we used PK for these experiments. For the BioNLP dataset, the improvements are not as significant. Further, as expected, the improvement is more prominent for smaller size of reference set~($M$). To optimize reference set $\bs{S}^R$ for $M=100$, it takes approximately 2 to 3 hours~(with $\beta=1000, \gamma=300$ in \algoref{alg:opt_ref}).

\noindent\subsubsection{Nonstationary Kernel Learning~(NSK):}

In \figref{fig:pubmed_ns} and \ref{fig:bionlp_ns}, we compare performance of non-stationary kernels, w.r.t. traditional stationary kernels~(M=100).
As proposed in \secref{sec:nst}, the idea is to extend a convolution kernel~(PK or GK) with non-stationarity-based binary parameters~(NS-PK or NS-GK), optimized using our MCMC procedure via maximizing the proposed MI approximation based objective~($\gamma=300$). For the PubMed45 dataset with PK, the advantage of NSK learning is more prominent, leading to high increase in recall~(7 pts), and a very small drop in precision~(1 pt). Compute time for learning the non-stationarity parameters in our KLSH-RF model is less than an hour.

\noindent\subsubsection{Compute Time:}

Compute times to train all the models are reported in \figref{fig:compute_time} for the BioNLP dataset; similar time scales apply for other datasets. We observe that our basic KLSH-RF model has a very low training cost, w.r.t. models like LSTM, KLSH-kNN, etc.~(similar analysis applies for inference cost). The extensions of KLSH-RF, KLSH-RF-RO and KLSH-RF-NS, are more expensive yet cheaper than LSTM and SVM.

\section{Related Work}
\label{sec:related_work}

Besides some related work mentioned in the previous sections, this section focuses on relevant state-of-the-art literature in more details.
				
\noindent\subsubsection{Other Hashing Techniques:}
In addition to  hashing techniques considered in this paper, other locality-sensitive hashing techniques~\cite{grauman2013learning,zhao2014locality,wang2017survey} are either not kernel based, or they are defined for specific kernels that are not applicable for hashing of NLP structures~\cite{raginsky2009locality}. In deep learning, hashcodes are used for similarity search but classification of objects~\cite{liu2016deep}.
			
\noindent\subsubsection{Hashcodes for Feature Compression:}
Binary hashing~(not KLSH) has been used as an approximate feature compression technique in order to reduce memory and computing costs  ~\cite{li2011hashing,mu2014hash}. Unlike prior approaches, this work proposes to use hashing as a representation learning (feature extraction) technique.

\noindent\subsubsection{Using Hashcodes in NLP:} In NLP, hashcodes were used only for similarity or nearest neighbor search for words/tokens in various NLP tasks~\cite{goyal2012fast,li2014two,shi2017fixed}; our work is the first to explore kernel-hashing  of various NLP structures, rather than just tokens.

\noindent\subsubsection{Weighting Substructures:} Our idea of skipping substructures, due to our principled approach of nonstationary kernels, is somewhat similar to   sub-structure mining algorithms~\cite{suzuki2006sequence,severyn2013fast}. Learning the weights    of sub-structures was recently proposed for regression problems, but not yet for  classification~\cite{beck2015learning}.

\noindent\subsubsection{Kernel Approximations: }
Besides the proposed model, there are other kernel-based scalable techniques in the literature, which  rely on approximation of a  kernel matrix or a kernel function~\cite{williams2001using,moschitti2006making,rahimi2008random,pighin2009efficient,zanzotto2012distributed,severyn2013fast,felix2016orthogonal}. However, those approaches are only used as  computationally efficient approximations of the traditional, computationally-expensive kernel-based classifiers; unlike those approaches, our method is not only  computationally more efficient but also yields considerable accuracy improvements.

\noindent\subsubsection{Nonstationary Kernels:}
Nonstationary kernels have been explored for modeling spatio-temporal environmental dynamics or time series relevant to health care, finance, etc, though expensive to learn due to a prohibitively large number of latent variables~\cite{paciorek2003nonstationary,snelson2003warped,assael2014heteroscedastic}. Ours is the first work proposing nonstationary convolution kernels for natural language modeling; the number of parameters is constant in our formulation, so highly efficient in contrast to the previous works.

\section{Conclusions}
In this paper we propose to use a well-known technique, kernelized locality-sensitive hashing (KLSH), in order to derive feature vectors from natural language structures. More specifically, we  propose to use random subspaces of KLSH codes for building a random forest of decision trees. We find this methodology particularly suitable for modeling natural language structures in supervised settings where there are significant mismatches between the training and the test conditions. 
Moreover we optimize a KLSH model in the context  of classification performed using a random forest, by maximizing an approximation of the mutual information between the KLSH codes (feature vectors) and the class labels. We apply the proposed approach to the difficult task of extracting information about bio-molecular interactions from the semantic or syntactic parsing of scientific papers. Experiments on a wide range of datasets demonstrate the considerable advantages of our method.

\section{Acknowledgments}
This work was sponsored by the DARPA Big Mechanism program (W911NF-14-1-0364). It is our pleasure to acknowledge fruitful discussions with Karen Hambardzumyan, 
Hrant Khachatrian, 
David Kale, 
Kevin Knight, 
Daniel Marcu,
Shrikanth Narayanan,
Michael Pust, 
Kyle Reing,
Xiang Ren,
and 
Gaurav Sukhatme. We are also grateful to anonymous reviewers for their valuable feedback.

{
\bibliographystyle{aaai}
\bibliography{references}{}
}

\appendix
\input{appendix.tex}

\end{document}

%% file: appendix.tex
\section{Dataset Statistics}
The number of valid/invalid extractions in each dataset is shown in \tabref{tab:data_stats}.

\begin{table*}[tp!]
\centering
%
\begin{tabular}{lll}
\toprule
\textbf{Datasets}
&\textbf{No. of Valid Extractions}&\textbf{No. of Invalid Extractions}
\\
\toprule
PubMed45&2,794&20,102\\
\midrule
BioNLP&6,527&34,958\\
\midrule
AIMed
&1,000&4,834\\
\midrule
BioInfer
&2,534&7,132\\
\toprule
\end{tabular}
\caption{
%
Dataset statistics: 
number of ``valid" \& ``invalid" extractions in each of the four datasets.
}
\label{tab:data_stats}
\end{table*}


\section{Nonstationarity of Convolution Kernels for NLP}
%
%
%
\begin{definition}[Stationary kernel~\cite{genton2001classes}]
A \emph{stationary} kernel, between vectors $\bs{x}_i, \bs{x}_j \in \mathbb{R}^d$, is the one which is translation invariant:
\begin{align*}
k(\bs{x}_i, \bs{x}_j) = k^S(\bs{x}_i - \bs{x}_j),
\end{align*}
\noindent that means, it depends only upon the lag vector between $\bs{x}_i$ and $\bs{x}_j$, and not the data points themselves.
\label{def:stationary_kernel}
\end{definition}
For NLP context, stationarity in convolution kernels is formalized as follows.
\begin{theorem}
A convolution kernel $K(.,.)$, a function of the kernel $k(.,.)$, is stationary if $k(.,.)$ is stationary. From a nonstationary $k^{NS}(.,.)$, the corresponding extension of $K(.,.)$, $K^{NS}(.,.)$, is also guaranteed to be a valid nonstationary convolution kernel.
\begin{proof}
Suppose we have a vocabulary set, $\{ l_1, \cdots, l_{p}, \cdots, l_{2p} \}$, and we randomly generate a set of discrete structures $\mathcal{X} = \{ X_1, \cdots,  X_N \}$, using $l_1, \cdots, l_p$. For kernel $k(.,.)$, that defines similarity between a pair of labels, consider a case of stationarity, $k(l_i,l_j) = k(l_{i+p}, l_{j}) = k(l_{i}, l_{j+p}); i,j \in \{1, \cdots, p\}$, where its value is invariant w.r.t. to the translation of a label $l_i$ to $l_{i+p}$. In the structures, replacing labels $l_1, \cdots, l_p$ with $l_{p+1}, \cdots, l_{2p}$ respectively, we obtain a set of new structures $\bar{\mathcal{X}} = \{ \bar{X}_1, \cdots, \bar{X}_N \}$. Using a convolution kernel $K(.,.)$, as a function of $k(.,.)$, we obtain same~(kernel) Gram matrix on the set $\bar{\mathcal{S}}$ as for $\mathcal{S}$. Thus $K(.,.)$ is also invariant w.r.t. the translation of structures set $\mathcal{S}$ to $\mathcal{\bar{S}}$, hence a stationary kernel~(Def. \ref{def:stationary_kernel}).
%
%
For establishing the nonstationarity property, following the above logic, if using $K^{NS}(.,.)$,
we obtain a (kernel) Gram matrix on the set $\bar{\mathcal{S}}$ that is different from the set $\mathcal{S}$.
Therefore $K^{NS}(.,.)$ is not invariant w.r.t. the translation of set $\mathcal{S}$ to $\mathcal{\bar{S}}$, hence a nonstationary kernel~(Def. \ref{def:stationary_kernel}).
\end{proof}
\end{theorem}

\section{Brief on MCMC Procedure for Optimizing Nonstationary Parameters}
Denoting all the nonstationary parameters as $\bs{\sigma}$, we set $\bs{\sigma} = \bs{1}$ as the first sample of MCMC. Now, for producing a new sample $\bs{\sigma}'$ from current sample $\bs{\sigma}$ in the Markov chain, we randomly pick one of the parameters and flip its binary value from 0 to 1 or vica versa. This new sample is accepted with probability: $A(\bs{\sigma}, \bs{\sigma}')
=
\min\left(1, 
\exp\left(\cI(\bs{C}': \bs{y})\right)/\exp\left(\cI(\bs{C}: \bs{y})\right)\right)$, with hashcodes $\bs{C}'$ and $\bs{C}$ computed using the kernel parameters samples $\bs{\sigma}'$ \& $\bs{\sigma}$, respectively. This procedure is performed for a fixed number of samples, and then the MCMC sample with highest $\bar{\cI}(.: .)$ is accepted.

\section{More Experiments}
\begin{figure*}[tp!]
\centering
%
\subfigure[
Hash Functions~(PubMed45)
]{
\includegraphics[
width=0.75\columnwidth]{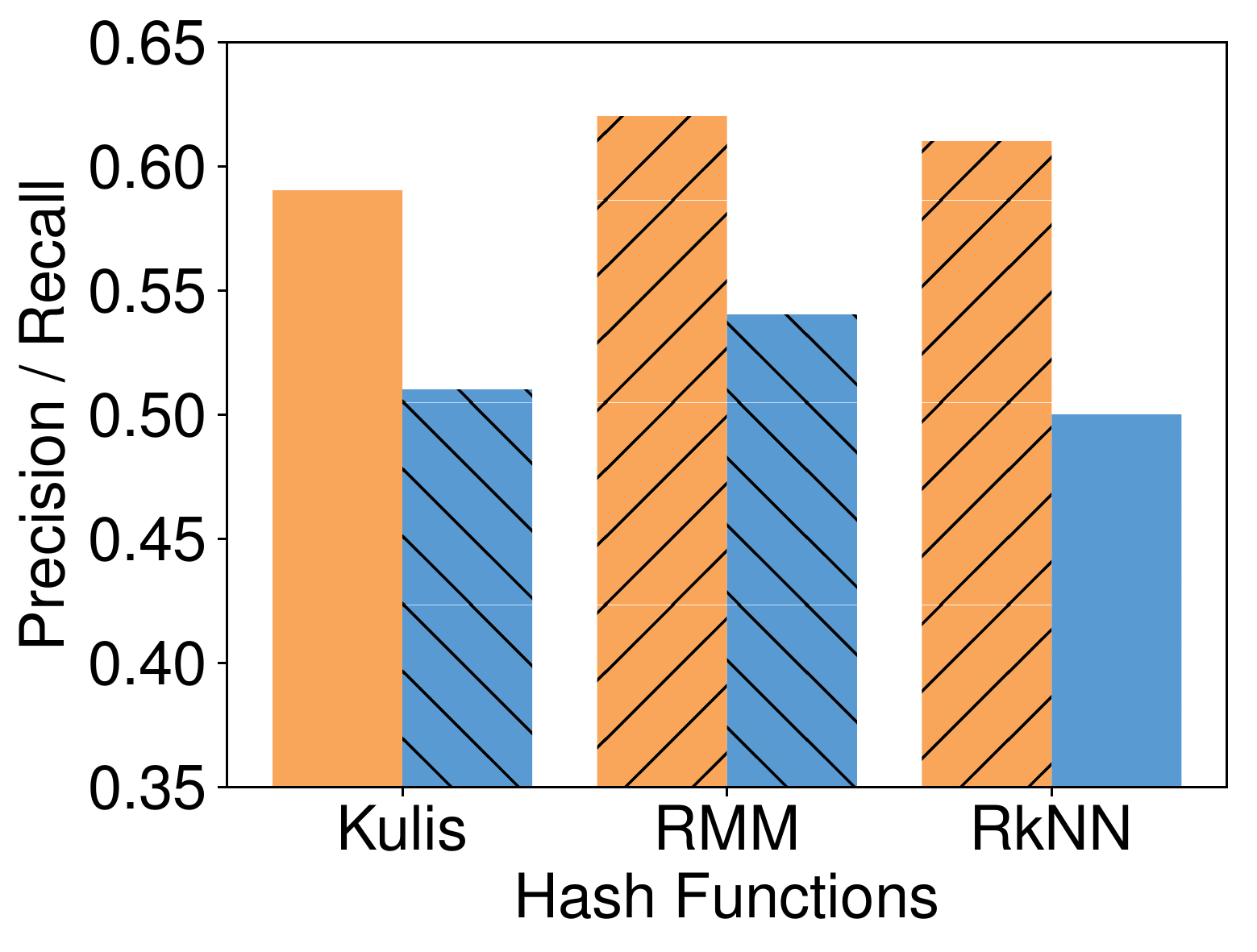}
\label{fig:pubmed_func}
}
\subfigure[
Hash Functions~(BioNLP)
]{
\includegraphics[
width=0.75\columnwidth]{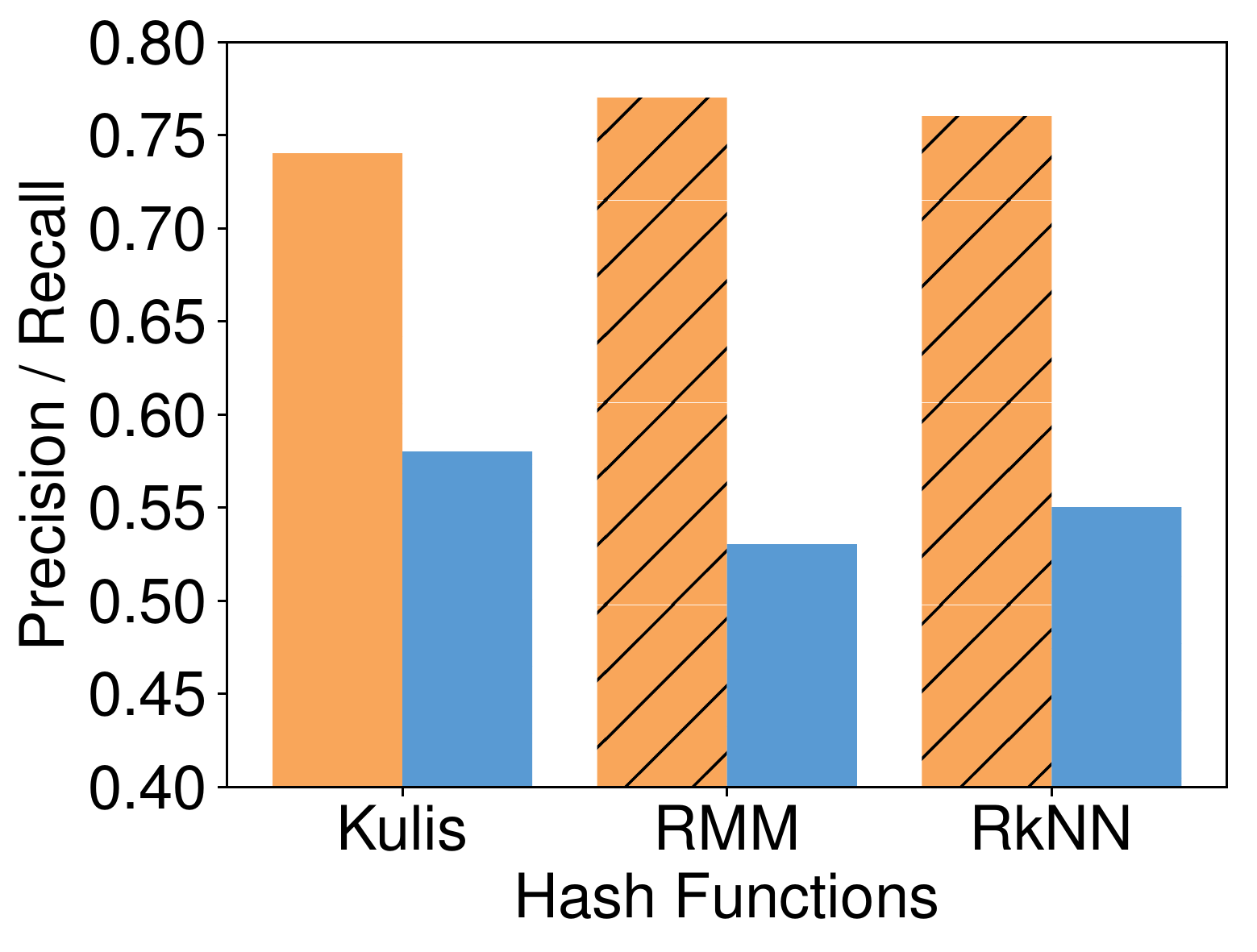}
\label{fig:bionlp_func}
}
%
%
%
\subfigure[
PubMed45- Sampling features
]{
\includegraphics[
width=0.75\columnwidth]{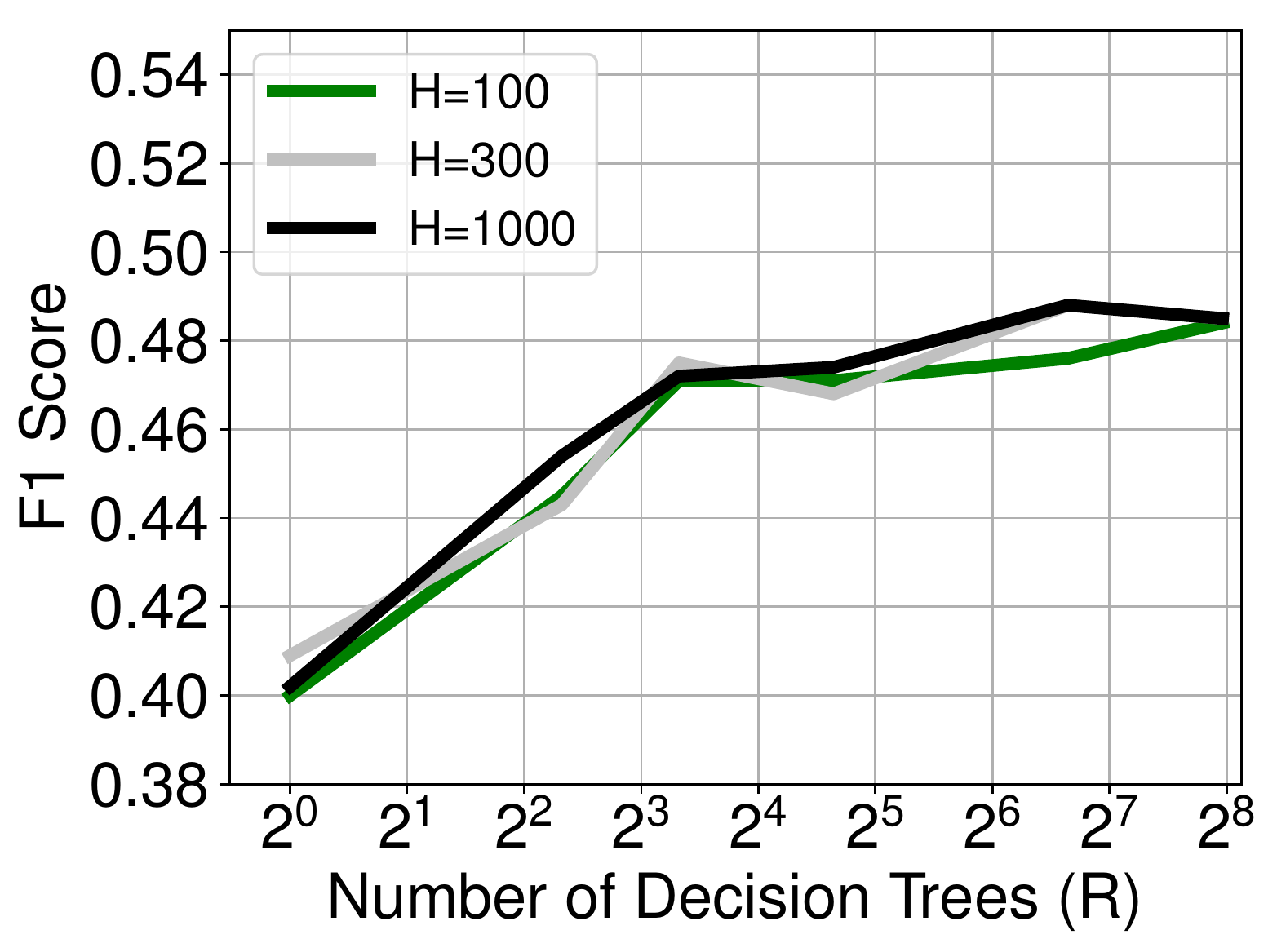}
\label{fig:subset}
}
%
%
%
\subfigure[
Varying $H, R$~(PubMed45)
%
]{
\includegraphics[
width=0.75\columnwidth
]{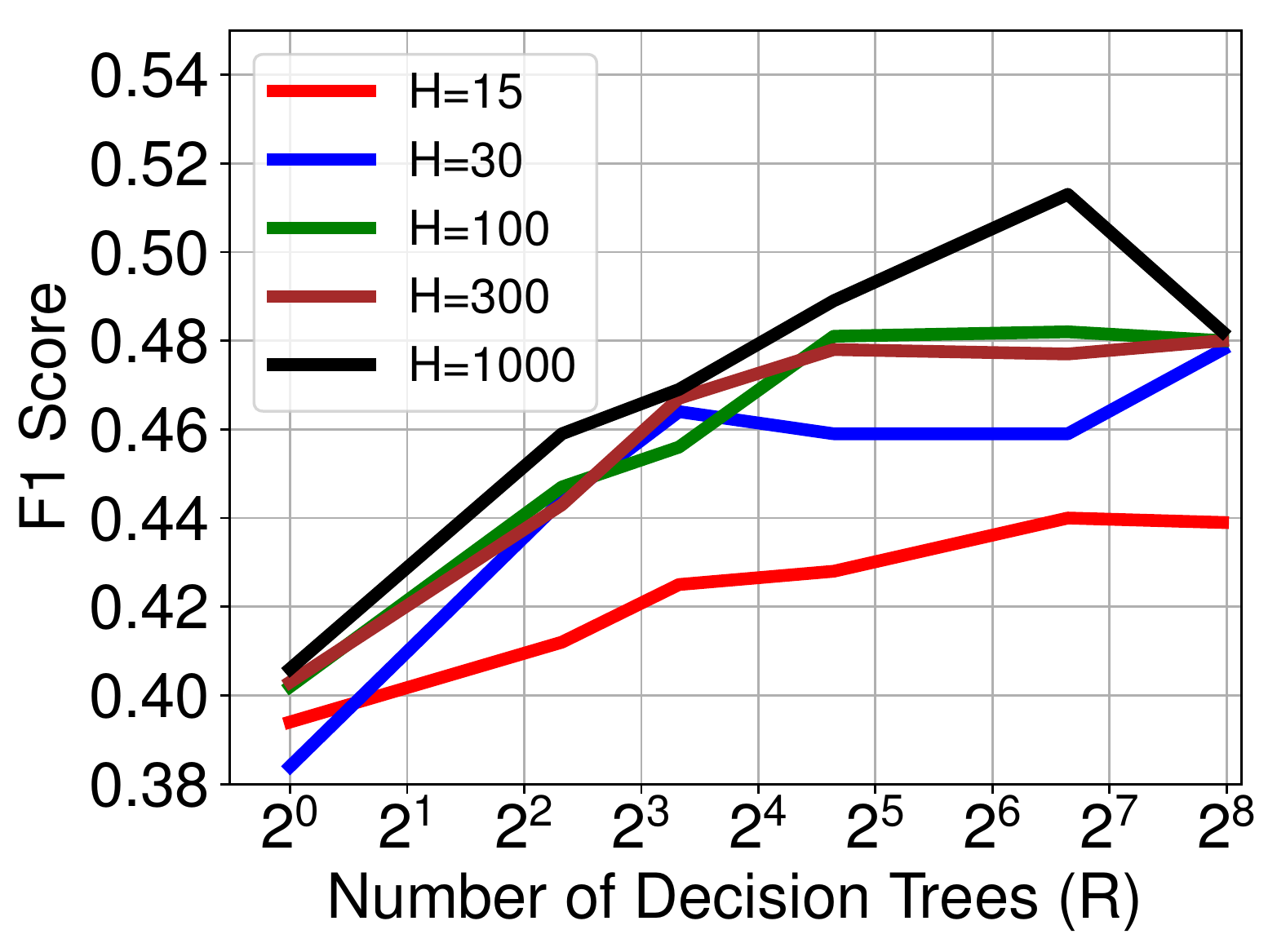}
\label{fig:bagging}
}
%
%
%
%
\caption{
%
Detailed Empirical Analysis of KLSH-RF model, using PubMed45 and BioNLP datasets. 
Here, orange and blue bars are for precision and recall numbers respectively.
%
}
\label{fig:klsh_rf_analysis}
\end{figure*}

\paragraph{Analyzing Hashing Parameters\\}
In \figref{fig:pubmed_func} and \ref{fig:bionlp_func}, we compare performance of all the three KLSH techniques with our model. For these experiments, $\alpha$ is fixed to value $2$. We found that Kulis is highly sensitive to the value of $\alpha$ in contrast to RMM and RkNN; accuracy numbers drop with Kulis for higher value of $\alpha$~(those results are not shown here).
%
    
For PubMed45 dataset, we also vary the parameters $R, H$~($\eta$=None \& $M=500$, using PK).
%
%
As we mentioned previously, for obtaining random subspaces of kernel-hashcodes, we can either use bagging~($\eta$=None), i.e. the random subset of training dataset~(with resampling), 
or explicitly take  a random subset of hashcode bits~($\eta=30$).
%
%
Here, in \figref{fig:subset} and \figref{fig:bagging}, we present results for both approaches, as two types of our KLSH-RF model, with PK. 
We can see that the gain in accuracy, is marginal, with an increase in the number of decision trees, after a minimal threshold. 
For a low value of $H$~(15, 30), the F1 score drops significantly.
In \ref{fig:subset}, we decrease $H$ down to value 100 only since the number of sampled hashcode bits~($\eta$) is 30. 
We also note that despite the high number of hashcode bits, classification accuracy improves only if we have a minimal number of 
decision trees.

\section{Convolution Kernels Expressions in Experiments}
\label{sec:convolution_kernels}
Convolution kernels belong to a class of kernels that compute similarity between discrete structures~\cite{haussler1999convolution,collins2002convolution}. In essence, convolution kernel similarity function $K(X_i, X_j)$ between two discrete structures $X_i$ and $X_j$, is defined in terms of function $k(.,.)$ that characterizes  similarity between a pair of tuples or labels. In the following, we desribe the exact expressions for convoution kernels used in our experiments while the proposed approaches are generically applicable for any convolution kernels operating on NLP structures.
     
\paragraph{Graph/Tree Kernels}
In \cite{zelenko2003kernel,garg2016extracting},  the kernel similarity between two trees $T_i$ and $T_j$ is defined as:
%
\begin{align*}
&
\csizeten
K(T_i, T_j) 
=
k(T_i.r, T_j.r)
(
k(T_i.r, T_j.r)
+
\!\!\!
\sum_{\bs{i}, \bs{j}: l(\bs{i})=l(\bs{j})}
\!\!
\lambda^{l(\bs{i})}
\\
&
\csizeten
\!\!
\sum_{s=1,\cdots,l(\bs{i})}
\!\!\!\!\!\!\!
K(T_i[\bs{i}[s]], T_j[\bs{j}[s]])
\!\!\!\!\!\!\!
\prod_{s=1,\cdots,l(\bs{i})}
\!\!\!\!\!\!\!
k(T_i[\bs{i}[s]].r, T_j[\bs{j}[s]].r)
),
\end{align*}
%
\noindent where $\bs{i}, \bs{j}$ are child subsequences under the root nodes $T_i.r, T_j.r$ and $\lambda \in (0,1)$ as shown above; 
$\bs{i}=(i_1, \cdots, i_l)$ and $T_i[\bs{i}[s]]$ are subtrees rooted at the $\bs{i}[s]_{th}$ child node of $T_i.r$. Note that we use exact same formulation as used in \cite{garg2016extracting}.
    
\paragraph{Path/Subsequence Kernels}
Let $S_i$ and $S_j$ be two sequences of tuples, in \cite{mooney2005subsequence} the kernel is defined as:

\begin{align*}
\csizeten
K(S_i, S_j)
=
\sum_{\boldsymbol{i}, \boldsymbol{j}: |\boldsymbol{i}|=|\boldsymbol{j}|}
\prod_{k=1}^{|\boldsymbol{i}|}
k(S_i(\bs{i}_k), S_j(\bs{j}_k))
\lambda^{l(\boldsymbol{i})+l(\boldsymbol{j})}.
\end{align*}

Here, $k(S_i(\bs{i}_k), S_j(\bs{j}_k))$ is the similarity between the $k_{th}$ tuples in the subsequences $\boldsymbol{i}$ and $\boldsymbol{j}$, of equal length; $l(.)$ is the actual length of a subsequence in the corresponding sequence, i.e., the difference between the end index and start index~(subsequences do not have to be contiguous); $\lambda \in (0, 1)$ is used to penalize the long subsequences.

\paragraph*{}
For both kernels above, dynamic programming is used for efficient computation.